\documentclass{article}

\PassOptionsToPackage{numbers, compress}{natbib}

\usepackage[preprint]{neurips_2023}

\usepackage[utf8]{inputenc} 
\usepackage[T1]{fontenc}    
\usepackage{hyperref}       
\usepackage{url}            
\usepackage{booktabs}       
\usepackage{amsfonts}       
\usepackage{amsmath}
\usepackage{amssymb}
\usepackage{algorithm}
\usepackage[pdftex]{graphicx}
\usepackage{utfsym}
\usepackage{algorithmic}
\usepackage{multirow}
\usepackage{nth}
\usepackage{nicefrac}       
\usepackage{microtype}      
\usepackage{xcolor}         %
\hypersetup{
    colorlinks  = true, 
    urlcolor    = blue, 
    linkcolor   = red, 
    citecolor   = green 
}

\DeclareMathOperator*{\argmin}{arg\,min}

\title{InstructP2P: Learning to Edit 3D Point Clouds with Text Instructions}

\author{
Jiale Xu\textsuperscript{\rm 1,2}\footnotemark[1] \quad
Xintao Wang\textsuperscript{\rm 1}\footnotemark[2] \quad
Yan-Pei Cao\textsuperscript{\rm 1} \quad
Weihao Cheng\textsuperscript{\rm 1} \\
\textbf{Ying Shan}\textsuperscript{\rm 1} \quad
\textbf{Shenghua Gao}\textsuperscript{\rm 2,3,4}\footnotemark[2] \\
\textsuperscript{\rm 1}ARC Lab, Tencent PCG \quad
\textsuperscript{\rm 2}ShanghaiTech University \\
\textsuperscript{\rm 3}Shanghai Engineering Research Center of Intelligent Vision and Imaging \\
\textsuperscript{\rm 4}Shanghai Engineering Research Center of Energy Efficient and Custom AI IC \\
}

\begin{document}

\maketitle

\renewcommand{\thefootnote}{\fnsymbol{footnote}}
\footnotetext[1]{Work done during an internship at ARC Lab, Tencent PCG.}
\footnotetext[2]{Corresponding Authors.}

\begin{abstract}

Enhancing AI systems to perform tasks following human instructions can significantly boost productivity. In this paper, we present InstructP2P, an end-to-end framework for 3D shape editing on point clouds, guided by high-level textual instructions. InstructP2P extends the capabilities of existing methods by synergizing the strengths of a text-conditioned point cloud diffusion model, Point-E, and powerful language models, enabling color and geometry editing using language instructions. To train InstructP2P, we introduce a new shape editing dataset, constructed by integrating a shape segmentation dataset, off-the-shelf shape programs, and diverse edit instructions generated by a large language model, ChatGPT. Our proposed method allows for editing both color and geometry of specific regions in a single forward pass, while leaving other regions unaffected. In our experiments, InstructP2P shows generalization capabilities, adapting to novel shape categories and instructions, despite being trained on a limited amount of data.

\end{abstract}

\section{Introduction}
\label{sec:intro}

Automatically editing 3D shapes is an important task beneficial to various applications such as computer graphics, computer-aided design, gaming, and animation. 
Conventional methods~\cite{botsch2007linear} focus on specific 3D manipulations using computational geometry algorithms, such as mesh deformation~\cite{sorkine2004laplacian, 10.1145/1073204.1073229, joshi2007harmonic} and surface subdivision~\cite{Loop1987, catmull1978recursively}. Although these methods offer precise control over shapes, the editing processes tend to be functionally simplistic, laborious, and incapable of comprehending high-level human instructions. In contrast, learning-based shape editing methods~\cite{achlioptas2018learning, tang2022neural, liu2021deepmetahandles, hao2020dualsdf, nakayama2023difffacto} can learn decomposed implicit representations directly from shape data, thus enabling smooth shape transitions and controlled manipulation at higher semantic levels. However, editing with implicit representations is neither handy nor intuitive, and these methods may suffer from limited generalization across unseen categories due to scarce training data, constraining their application in real-world scenarios. Moreover, existing shape editing techniques often encounter difficulties when attempting to perform color and geometry editing simultaneously.

Employing natural language instructions to manipulate 3D shapes offers a more user-friendly and intuitive way of editing, which can significantly improve productivity.
With the rise of vision-language models pre-trained on large-scale text-image datasets, such as CLIP~\cite{radford2021learning}, recent studies~\cite{Michel_2022_CVPR, chen2022tango, ma2023xmesh, Gao_2023_SIGGRAPH} have explored optimization-based mesh stylization by iteratively maximizing the similarity between multi-view shape renderings and a text prompt. Despite the promising results, these methods are limited to generating simple vertex deformations and color variations, and the optimization process can be time-consuming. More recently, \citet{achlioptas2022changeIt3D} propose to train a shape auto-encoder and a neural listener that distinguishes the target shape from a distractor with a text description first, and then train a shape editor module to edit shapes within the latent space in a way that is both consistent with the language instruction and also minimal. However, their method is limited to specific shape categories and only allows for geometry editing.

To address the limitations of existing methods, we draw inspiration from the recent work on instruction-guided image editing~\cite{brooks2022instructpix2pix} and present InstructP2P, a novel end-to-end 3D shape editing framework. InstructP2P aims to automate the editing process by employing high-level textual instructions, thereby enabling users to effectively manipulate complex 3D objects. 
Considering that instruction-based point cloud manipulation requires a comprehensive understanding of language semantics, geometric structures, visual appearance, and the correspondence between language, shape, and texture, we propose to collect a dataset encompassing these features to facilitate the training of InstructP2P. Specifically, we leverage a shape segmentation dataset, \emph{i.e.}, PartNet~\cite{Mo_2019_CVPR}, along with the shape programs of three categories provided by~\citet{pearl2022geocode} to produce color and geometry editing instances. Additionally, we harness the power of a large language model (LLM), \emph{i.e.}, ChatGPT~\cite{OpenAI2022ChatGPT}, to generate a plethora of edit instructions. 

By training on a compilation of shape editing examples and the corresponding edit instructions, we transform a text-conditioned point cloud diffusion model, \emph{i.e.}, Point-E~\cite{nichol2022point}, into an instruction-conditioned shape editor, which is capable of editing both the color and geometry of a designated region in a single forward pass by following the text instruction, while preserving other regions untouched. Thanks to the rich 3D priors and impressive shape generation ability of Point-E, InstructP2P also exhibits a degree of generalization to previously unseen shape categories and instructions, even though it has only been trained on a limited set of editing instances.

The main contributions of our paper can be summarized as follows: 1) We introduce InstructP2P, the first general instruction-guided 3D shape and color editing framework that marries the strengths of text-conditioned point cloud diffusion models with intuitive instructional guidance, enabling  effective color and geometry editing. 2) We present a new shape editing dataset built upon an existing shape segmentation dataset and off-the-shelf shape programs, encompassing a diverse collection of shape editing examples and the corresponding edit instructions generated by a powerful LLM. 3) Our proposed method, InstructP2P, exhibits generalization capabilities to unseen shape categories and edit instructions, expanding its range of real-world applications and laying the foundation for future advancements in text-guided 3D shape editing.

\begin{figure}[t]
  \centering
  \vspace{-1cm}
  \includegraphics[width=0.98\linewidth]{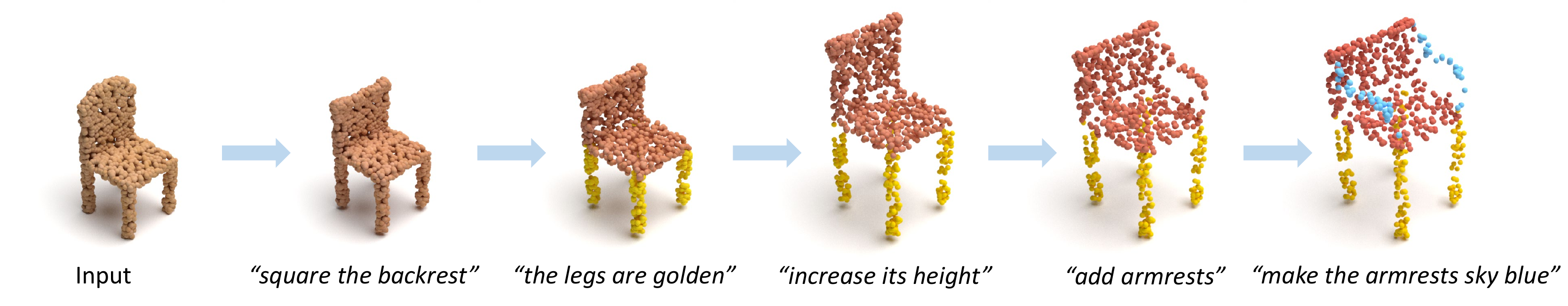}
  \vspace{-4mm}
  \caption{Instruction-guided 3D shape editing. Given a 3D shape represented by a point cloud, our InstructP2P model can perform accurate color and geometry editing following text instructions.}
  \label{fig:serialized}
  \vspace{-6mm}
\end{figure}

\section{Related Work}
\label{sec:related}

\paragraph{3D Shape Editing.}
3D shape editing is challenging and requires complex shape understanding. 
Traditional methods use explicit deformations~\cite{botsch2007linear, sorkine2004laplacian, 10.1145/1073204.1073229, joshi2007harmonic}, while learning-based methods~\cite{achlioptas2018learning, tang2022neural, liu2021deepmetahandles, hao2020dualsdf, nakayama2023difffacto} employ lower-dimensional implicit representations~\cite{park2019deepsdf, mescheder2019occupancy} to enable manipulation at higher semantic levels.
Language-driven 3D shape editing has been studied to achieve more intuitive manipulation. Leveraging powerful vision-language models like CLIP~\cite{radford2021learning}, a series of works~\cite{Michel_2022_CVPR, chen2022tango, ma2023xmesh, Gao_2023_SIGGRAPH} generate mesh vertex deformations and colors by continuously optimizing the CLIP similarity between the rendered images and a text prompt. 
ChangeIt3D~\cite{achlioptas2022changeIt3D} builds a shape auto-encoder for presenting shapes in a latent space, and a neural listener that can distinguish target shapes from distractor shapes based on text descriptions. It then uses the two modules to train a latent shape editor that edits shapes according to input instructions with minimal changes.

\paragraph{Text-driven 2D Image Editing.}
The vision-language alignment ability of CLIP can be applied in text-driven image manipulation by altering the image content directly~\cite{bar2022text2live, lee2023shape} or optimizing the latent code of a pre-trained Generative Adversarial Network (GAN)~\cite{goodfellow2020generative, Karras_2020_CVPR, patashnik2021styleclip, lyu2023deltaedit, crowson2022vqgan, zheng2022bridging}. Recent advances in text-to-image (T2I) diffusion models~\cite{ho2020denoising, song2021denoising, Rombach_2022_CVPR, saharia2022photorealistic} trained on internet-scale text-image datasets provide a new paradigm for text-driven image editing~\cite{Kim_2022_CVPR, meng2022sdedit}. Recent works~\cite{kawar2023imagic, hertz2022prompt, mokady2022null, couairon2023diffedit, zhang2022sine} have leveraged T2I diffusion models to conduct free-style image manipulation in a diffusion-denoising scheme, allowing for greater control and flexibility. \citet{brooks2022instructpix2pix} further presents an instruction-guided image editing model that is fine-tuned from a T2I diffusion model using large-scale synthetic image editing pairs associated with edit instructions. 
Following~\cite{brooks2022instructpix2pix}, some methods have also explored instruction-guided image inpainting~\cite{yildirim2023inst} and neural radiance field stylization~\cite{haque2023instruct, kamata2023instruct}.

\paragraph{3D Diffusion Models.}
Diffusion models, gaining popularity for their generation performance and training stability, have been utilized in 3D shape generation for point clouds~\cite{Zhou_2021_ICCV, Luo_2021_CVPR, melaskyriazi2023projection, tyszkiewicz2023gecco}, meshes~\cite{gupta20233dgen, Liu2023MeshDiffusion, lyu2023controllable}, and implicit fields~\cite{shue20223d, zeng2022lion, li2022diffusion, cheng2022sdfusion, hu2023neural, li20233dqd, erkoç2023hyperdiffusion}. Despite the promising shape generation performance, these methods are often trained on limited shape categories and are hard to generalize. Recently, OpenAI introduced Point-E~\cite{nichol2022point}, a conditional point cloud diffusion model trained on millions of 3D models. Thanks to the large-scale training data, Point-E can generate colored point clouds from complex text or image prompts and exhibits great generalization ability across many shape categories. Our instruction-guided shape editing model is built upon Point-E to enjoy its rich internal 3D priors.

\begin{figure}[t]
  \centering
  \vspace{-0.3cm}
  \includegraphics[width=0.98\linewidth]{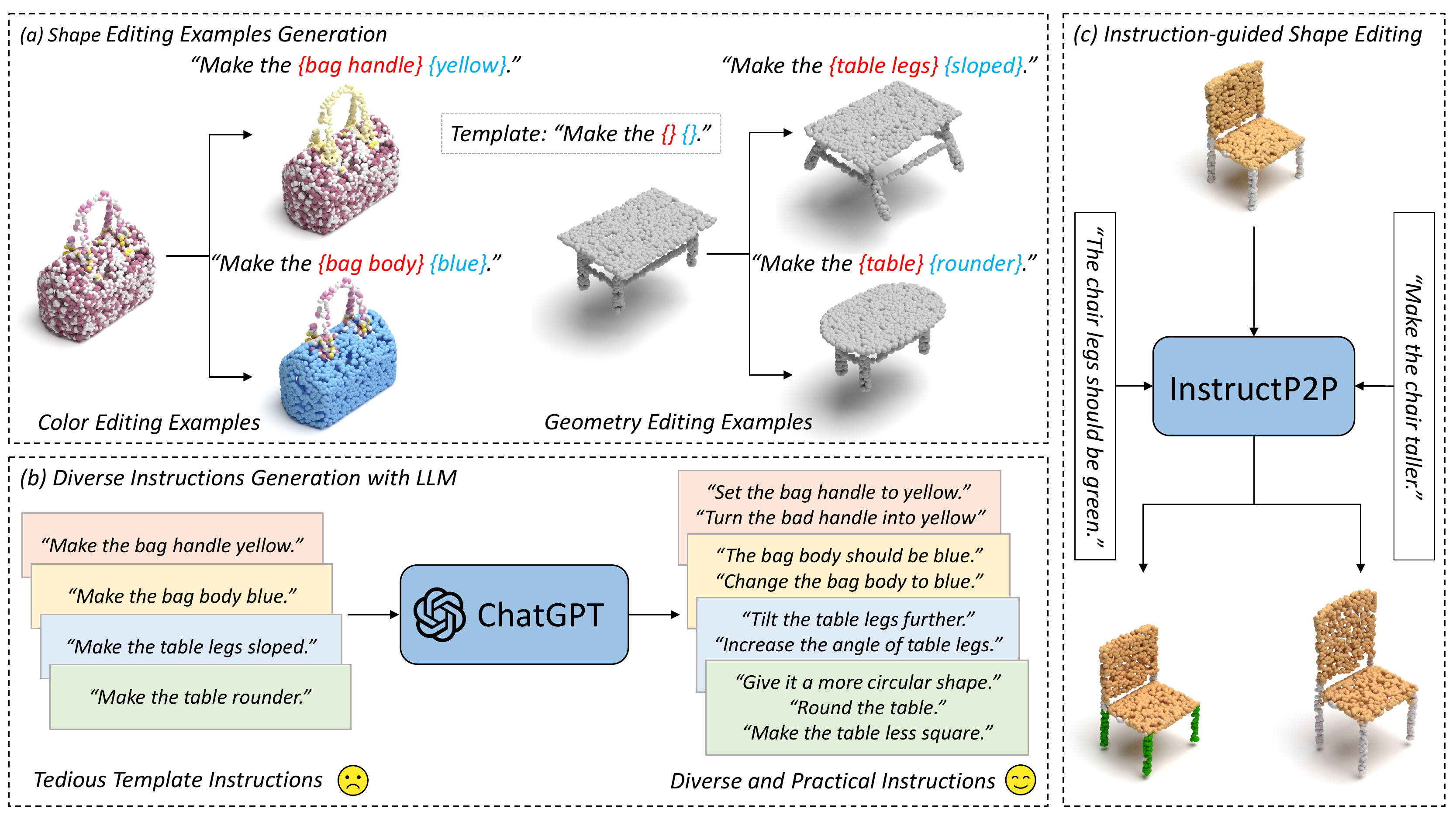}
  \caption{
  Overview. (a)
  Color and geometry editing examples. We utilize a text template \emph{``Make the \textcolor{red}{\{\}} \textcolor{cyan}{\{\}}''} to create simple edit instructions for the editing examples pairs, where \textcolor{red}{\{\}} denotes the editing objective and \textcolor{cyan}{\{\}} is a description of the editing. 
  (b) To improve the robustness of the trained model and simulate a more realistic user experience, we leverage ChatGPT to diversify the edit instructions. (c) After training InstructP2P on the dataset, we can edit the color and geometry of an input point cloud following text instructions in a single forward pass.}
  \label{fig:pipeline}
  \vspace{-4mm}
\end{figure}

\paragraph{Large Language Models.}
Advancements in large language models (LLMs) have greatly impacted natural language understanding. Earlier models, such as word2vec~\cite{mikolov2013distributed} and GloVe~\cite{pennington2014glove}, introduced continuous word embeddings, while transformer architectures, like GPT~\cite{radford2018improving} and BERT~\cite{DBLP:conf/naacl/DevlinCLT19}, utilized self-attention for improved context representation. Recent LLMs, including GPT-3~\cite{brown2020language}, PaLM~\cite{chowdhery2022palm}, ChatGPT~\cite{OpenAI2022ChatGPT}, and GPT-4~\cite{openai2023gpt4}, showcased their potential across various sophisticated tasks such as question-answering, summarization, translation, and code completion. These new models showcased unprecedented capabilities thanks to their massive scale and pre-training techniques, highlighting the potential of LLMs in a vast spectrum of applications.

\section{Method}
\label{sec:method}

Given a 3D shape represented by a colored point cloud $\mathbf{P}\in \mathbb{R}^{N\times 6}$ and a text instruction $\mathbf{t}$ describing the desired color/geometry modification, we aim to synthesize a new point cloud $\mathbf{P}'\in \mathbb{R}^{N\times 6}$ where the color or geometry is accurately edited according to $\mathbf{t}$. We present InstructP2P, an end-to-end model that generates edited shapes conditioned on given shapes and text instructions. The architecture of InstructP2P is based on Point-E, a powerful text-to-3D diffusion model. We directly take the pre-trained weights from Point-E and fine-tune InstructP2P with a collection of editing examples where each example includes a source point cloud, a target point cloud, and an edit instruction.

In this section, we first outline background knowledge (Section~\ref{sec:backgroud}). We then detail our data generation pipeline with automatic editing examples and ChatGPT-generated instructions (Section~\ref{sec:data}). Finally, we describe the model architecture and training procedures of InstructP2P (Section~\ref{sec:instructp2p}).

\subsection{Preliminaries}
\label{sec:backgroud}

\subsubsection{Text-conditioned Point Cloud Diffusion Model}

Diffusion models~\cite{ho2020denoising, song2021denoising} learn to gradually denoise pure Gaussian noise into data samples. Given a data sample $\mathbf{x}_0 \sim q\left(\mathbf{x}_0\right)$ and a noise schedule $0<\beta_1<\cdots<\beta_T<1$, the forward diffusion process $q$ gradually adds noise to $\mathbf{x}_0$ at each time step $t$: $q\left(\mathbf{x}_t|\mathbf{x}_{t-1}\right)=\mathcal{N}\left(\mathbf{x}_t ; \sqrt{1-\beta_t} \mathbf{x}_{t-1}, \beta_t \mathbf{I}\right)$. Let $\alpha_t=1-\beta_t$ and $\bar{\alpha}_t=\prod_{i=1}^t \alpha_i$, it can be achieved by sampling $\epsilon \sim \mathcal{N}(\mathbf{0}, \mathbf{I})$ and then setting $\mathbf{x}_t=\sqrt{\bar{\alpha}_t} \mathbf{x}_0+\sqrt{1-\bar{\alpha}_t} \mathbf{\epsilon}$. To reverse this process, a noise prediction network $\epsilon_{\theta}$ is trained to estimate the added noise at each timestep $t$ using the following loss function:

\begin{equation}
\mathcal{L}=\mathbb{E}_{\mathbf{x}_0, t, \epsilon_t}\left\|\epsilon_t-\epsilon_\theta\left(\sqrt{\bar{\alpha}_t} \mathbf{x}_0+\sqrt{1-\bar{\alpha}_t} \epsilon_t; t\right)\right\|_2^2
\label{eq:diffusion_loss}
\end{equation}

After training, we can sample a random Gaussian noise $\mathbf{x}_T\sim \mathcal{N}(\mathbf{0}, \mathbf{I})$ and then gradually denoise it to end up with a sample from the real distribution $\mathbf{x}_0$. Conditional diffusion models further add an additional condition $y$ to the denoising network $\epsilon_{\theta}$, such as a text or an image prompt, and the training objective becomes:

\begin{equation}
\mathcal{L}=\mathbb{E}_{\mathbf{x}_0, t, \epsilon_t, y}\left\|\epsilon_t-\epsilon_\theta\left(\sqrt{\bar{\alpha}_t} \mathbf{x}_0+\sqrt{1-\bar{\alpha}_t} \epsilon_t; t, y\right)\right\|_2^2
\label{eq:diffusion_loss_cond}
\end{equation}

Point-E~\cite{nichol2022point} is a conditional point cloud diffusion model released by OpenAI, which is trained on millions of 3D models associated with text descriptions. Given a text or an image prompt, Point-E uses its CLIP embedding as the condition of $\epsilon_{\theta}$ and gradually denoise a pure Gaussian noise into a colored point cloud. Point-E provides several model variants, which are trained under different settings: 1) \textit{40M-textvec}: a model conditioned on CLIP text embeddings of text descriptions, 2) \textit{40M-imagevec}: a model conditioned on CLIP image embeddings of rendered images, and 3) \textit{40M/300M/1B}: models that are condition on grid-features of rendered images encoded by the CLIP ViT image encoder, they share the same architecture but differ in parameter scales. Our InstructP2P is fine-tuned from the \textit{40M-textvec} variant of Point-E, due to its inherent ability to deal with language input.

\subsubsection{InstructPix2Pix}


\citet{brooks2022instructpix2pix} propose InstructPix2Pix, an end-to-end image editing framework that can follow human instructions. It leverages the power of an advanced large-scale text-to-image diffusion model, \emph{i.e.}, Stable Diffusion~\cite{Rombach_2022_CVPR}, and fine-tune it on a dataset containing 454K image editing examples and the corresponding instructions. To generate the dataset, \citet{brooks2022instructpix2pix} first manually create 700 \emph{<source caption, target caption, edit instruction>} triplets, then leverage them to fine-tune a GPT-3~\cite{brown2020language} language model to make it generate a larger amount (454K) of triplets. To transform the \emph{<source caption, target caption>} pairs into \emph{<source image, target image>} pairs, they use Prompt-to-Prompt~\cite{hertz2022prompt}, a recent method aimed at encouraging multiple generations from a text-to-image diffusion model to be similar. Finally, the \emph{<source image, target image, edit instruction>} triplets are used to fine-tune Stable Diffusion into the InstructPix2Pix model.

\subsection{Dataset Generation}
\label{sec:data}

To give users more control, our method aims to handle both part-level and global-level shape editing, as well as color and geometry manipulations simultaneously. Inspired by InstructPix2Pix, we propose to leverage the text-to-shape generation ability of Point-E and fine-tune it into an instruction-guided shape editing model. However, the training requires a shape editing dataset containing color and geometry editing examples associated with edit instructions, which is challenging to collect since there is no automatic tool like Prompt-to-Prompt to create precise shape pairs from caption pairs in the 3D community. We now illustrate how we generate such editing examples and edit instructions.

\subsubsection{Generating Color Editing Examples}

We exploit a 3D shape dataset with part-level annotations, \emph{i.e.}, PartNet~\cite{Mo_2019_CVPR}, to generate the color editing examples. PartNet consists of 27K 3D models covering 24 object categories annotated with fine-grained and hierarchical 3D part information. In this work, we utilize the first-level part annotations, please refer to the supplementary material for more details.

Given a colored point cloud $\mathbf{P} \in \mathbb{R}^{N \times 6}$ in PartNet with $K$ part annotations $\{\mathbf{p}_j\}_{j=1}^K$ ($\mathbf{P} = \mathbf{p}_1 \cup \cdots \cup \mathbf{p}_K$), for each part $\mathbf{p}_j$, we randomly assign a new color to the points. Denoting the edited part as $\mathbf{p}'_j$, we now obtain an edited point cloud $\mathbf{P}' = \mathbf{p}_1 \cup \cdots \cup \mathbf{p}'_j \cup \cdots \cup \mathbf{p}_K \in \mathbb{R}^{N \times 6}$. Meanwhile, we also generate an edit instruction $\mathbf{t}$ with a pre-defined text template like \emph{``make the \{part name\} \{color name\}''} and form an edit triplet <$\mathbf{P}, \mathbf{P}', \mathbf{t}$>. For example, if we assign a blue color to the ``legs'' of a chair, the edit instruction would be \emph{``make the chair legs blue''}. Following this scheme, we generate a part-level color editing dataset of \emph{<source point cloud, target point cloud, edit instruction>} triplets.

\subsubsection{Generating Geometry Editing Examples}

Our framework seeks to handle geometry editing tasks like part addition, deletion, and deformation. Acquiring a high-quality part-level dataset is challenging due to the absence of effective automatic tools for these operations.

Recently, \citet{pearl2022geocode} introduced GeoCode, which maps input shapes to an editable parameter space for novel shape assembly using Blender's~\cite{blender} Geometry Nodes. They created three programs for chairs, vases, and tables that decompose shapes into components, featuring human-interpretable parameters controlling properties. More details can be found in their paper and supplementary material.

We then present a two-step geometry editing dataset generation pipeline using these shape programs. In the first step, we assign random parameters to the shape programs and generate a large set of diverse shapes (triangle meshes) $\{S_i\}_{i=1}^M$. Next, for each mesh $S_i$, we iterate over its editable parameters and randomly alter each parameter to obtain a set of edited meshes $\{S_i^j\}_{j=1}^L$, where $L$ is the number of editable parameters. The edited shape $S_i^j$ shares exactly the same parameters with $S_i$ except for the $j$-th parameter. For each of the three shape categories, we define a unique set of editable shape parameters. For example, for the "chair" category, the editable parameters include leg length, backrest curvature, indicator of having armrests, etc. We give more details in the supplementary material.

When altering a parameter, we generate an associated edit instruction $\mathbf{t}$ from a pre-defined textual template according to whether the parameter is increased or decreased. For example, if the length of the chair leg is increased by the control parameter, the edit instruction would be \emph{``make the chair legs longer''}; otherwise, it would be \emph{``make the chair legs shorter''}. After iterating over every shape $S_i$ and every editable parameter of $S_i$, we obtain a geometry editing dataset consists of triplets $\{<S_k, S'_k, \mathbf{t}_k>\}_{k=1}^{ML}$. We then uniformly sample the source and target point clouds $\mathbf{P}_k, \mathbf{P}'_k \in \mathbb{R}^{N \times 6}$ from the surface of $S_k$ and $S'_k$, respectively, and get the final part-level geometry editing dataset $\{<\mathbf{P}_k, \mathbf{P}'_k, \mathbf{t}_k>\}_{k=1}^{ML}$. To be noted, the generated meshes are textureless, thus we assign a default gray color $(0.5, 0.5, 0.5)$ to all sampled point clouds.

\subsubsection{Enriching Edit Instructions with ChatGPT}

Although we have generated the associated edit instructions when generating the color/geometry editing examples, these instructions are created from hand-crafted templates and lack of diversity. Considering the powerful descriptive ability of large language models (LLMs), we leverage ChatGPT~\cite{OpenAI2022ChatGPT} to generate a more diverse set of edit instructions. For each existing instruction, we ask ChatGPT to rewrite it into three different shape edit instructions without altering the meaning. We give the full prompt we use in the supplementary material.

The usage of LLMs is highly advantageous. Firstly, ChatGPT can generate highly diverse and contextually relevant edit instructions, enhancing the robustness of the trained model. Secondly, the extended instructions can capture more complex and highly-specific shape edit instructions, which may otherwise be challenging to create manually using fixed templates. Lastly, utilizing a powerful LLM ensures the generated instructions reflect human-like expressing preferences, thus simulating a more realistic user experience and increasing the applicability in real-world scenarios.

\begin{figure}[t]
  \centering
  \includegraphics[width=0.98\linewidth]{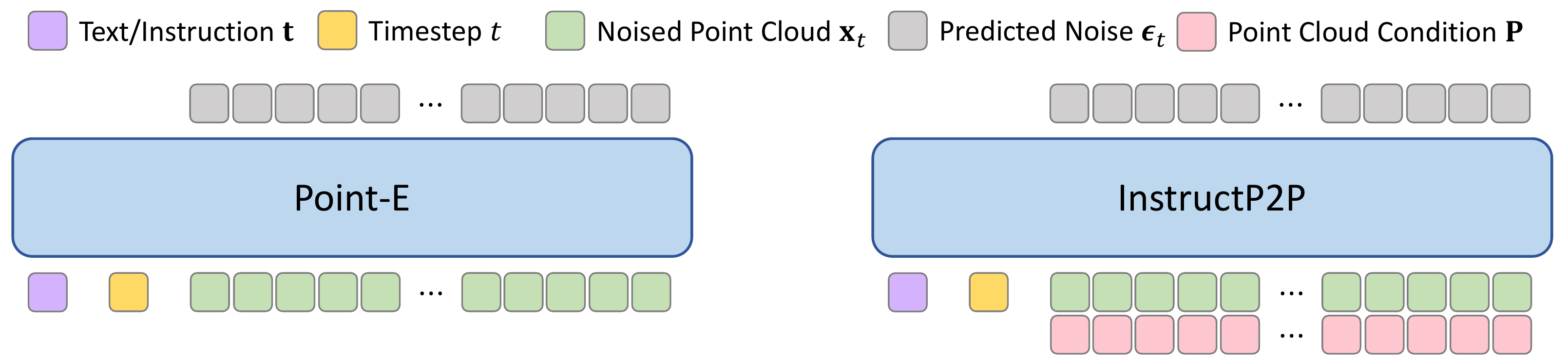}
  \vspace{-10pt}
  \caption{The transformer-based model architectures of Point-E and InstructP2P. The input of Point-E includes three parts: a token of the CLIP embedding of the text prompt, a token of the timestep $t$, and $N$ tokens of the noised point cloud $\mathbf{x}_t$. Our InstructP2P model replaces the text prompt with the edit instruction and concatenates the input point cloud $\mathbf{P}$ with $\mathbf{x}_t$ as an additional condition.}
  \vspace{-10pt}
  \label{fig:architecture}
\end{figure}

\subsection{InstructP2P}
\label{sec:instructp2p}

With the generated color and geometry editing examples paired with diverse text instructions, we can now train an end-to-end instruction-guided shape editing model.

\subsubsection{Model Architecture}
\label{sec:model}

The model architecture of InstructP2P is based on Point-E~\cite{nichol2022point}, which exploits a transformer-based model as the denoising network $\epsilon_{\theta}$. As shown in Figure~\ref{fig:architecture}, the original model maps the noised point cloud $\mathbf{x}_t \in \mathbb{R}^{N\times 6}$, the timestep $t\in \mathbb{R}$, and the CLIP text embedding $\mathbf{e}\in\mathbb{R}^{768}$ into $D$-dimensional tokens with separate linear layers and feeds them into the transformer to predict the added noise $\epsilon_{t}$ at timestep $t$. 

To transform the model into an end-to-end shape editing model, we add an additional point cloud condition $\mathbf{P} \in \mathbb{R}^{N \times 6}$ to $\epsilon_{\theta}$ by concatenating $\mathbf{x}_t$ and $\mathbf{P}$ and adding 6 extra input channels to the linear layer that maps $\mathbf{x}_t$ to transformer tokens. We initialize the weights of our model with the \emph{40M-textvec} checkpoint of Point-E, and the weights that operate on the newly added 6 input channels are initialized as 0.


\subsubsection{Training Details}
\label{sec:training}

The training objective of InstructP2P is the diffusion loss in Equation~\ref{eq:diffusion_loss_cond} with an extra point cloud condition, denoted as:
\begin{equation}
\mathcal{L}=\mathbb{E}_{\mathbf{P}', t, \epsilon_t, \mathbf{t}, \mathbf{P}}\left\|\epsilon_t-\epsilon_\theta\left(\sqrt{\bar{\alpha}_t} \mathbf{P}'+\sqrt{1-\bar{\alpha}_t} \epsilon_t; t, E_T(\mathbf{t}), \mathbf{P}\right)\right\|_2^2, 
\label{eq:diffusion_loss_p2p}
\end{equation}
where $\mathbf{t}$ represents the text instruction and $E_T$ signifies the CLIP text encoder. $\mathbf{P}'$ is the ground truth target point cloud corresponding to $\mathbf{P}$ which denotes the real data sample $\mathbf{x}_0$ in Equation~\ref{eq:diffusion_loss_cond}. The learning rate is set to $10^{-5}$, which is $10\times$ smaller than the original learning rate of Point-E. The point coordinates and RGB values in the dataset are both normalized to $[-1, 1]$ before feeding them into the model. We train InstructP2P on our shape editing dataset for 200K steps with a batch size of 64 using two NVIDIA A100 GPUs. The training takes about 48 hours.

\paragraph{Point Cloud Alignment.}

The transformer architecture of Point-E does not employ positional encodings for the input noise, thus ensuring permutation-invariant point cloud generation. However, since the output order is tied to the input order and we take the source point cloud $\mathbf{P}$ as an additional input, we need to make an alignment between the source point cloud $\mathbf{P}$ and the target point cloud $\mathbf{P}'$. For the color editing dataset, this issue does not exist since we only change the colors of a point subset and the source and target points strictly share the same order. However, the point cloud pairs in the geometry editing dataset are sampled from two different meshes and are not aligned in 3D space.

To align $\mathbf{P}$ and $\mathbf{P}'$, we first run the Iterative Closest Point (ICP) registration algorithm~\cite{121791} and apply the output $SE(3)$ transformation to $\mathbf{P}'$ to make them aligned globally. Then we construct a bijection $f: \mathbf{P} \to \mathbf{P}' $ by minimizing the sum of the squared Euclidean distances of all matched point pairs:
\begin{equation}
    f = \argmin_f \sum_{j=1}^N\|\mathbf{p}_j - f(\mathbf{p}_j)\|_2^2, 
    \label{eq:alignment}
\end{equation}
where $\mathbf{p}_j\in \mathbf{P}, f(\mathbf{p}_j) \in \mathbf{P}'$. This is a classical linear sum assignment problem and can be efficiently implemented with the SciPy~\cite{2020SciPy-NMeth} library. With the bijection $f$, we can permute $\mathbf{P}'$ so that $\mathbf{p}_j$ and $f(\mathbf{p}_j)$ share the same order. Intuitively, our model tries to learn the correct point displacements for a given instruction from the geometry editing examples, and this alignment operation makes the point displacements ``minimal'', thus easing the model learning.

\section{Experiments}
\label{sec:exp}

In this section, we first evaluate the efficacy of our proposed InstructP2P on the instruction-guided 3D shape editing task, both qualitatively (Section~\ref{sec:exp_qual}) and quantitatively (Section~\ref{sec:exp_quan}). Subsequently, we conduct ablation studies to assess the effectiveness of critical components of our framework (Section~\ref{sec:exp_ablation}). 

\subsection{Qualitative Evaluation}
\label{sec:exp_qual}

\paragraph{Baseline.}
Since there is no text-guided point cloud editing method that is publicly available, we draw inspiration from recent diffusion-model-based image editing methods and build a baseline method ourselves for comparison. Similar to Prompt-to-Prompt~\cite{hertz2022prompt}, our baseline requires three inputs: the input point cloud, a source prompt $\mathbf{t}_{\text{src}}$ captioning the input, and a target prompt $\mathbf{t}_{\text{tgt}}$ describing the edited point cloud. For example, if we want to edit the backrest of a chair into red color, then the source and target prompt would be \emph{``a chair''} and \emph{``a chair with red backrest''}, respectively. Given an input point cloud $\mathbf{P}\in\mathbb{R}^{N\times 6}$, we first run DDIM~\cite{song2021denoising} inversion using $\mathbf{t}_{\text{src}}$ as condition to inverse $\mathbf{P}$ into a noise $\mathbf{z}\in\mathbb{R}^{N\times 6}$, and then run the denoising sampling process from $\mathbf{z}$ using $\mathbf{t}_{\text{tgt}}$ as condition to generate the edited point cloud. Additionally, we leverage a ``strength'' hyper-parameter $s\in (0, 1.0]$ to control how many DDIM inversion steps we run. When $s=1.0$, $\mathbf{P}$ is inverted into a pure Gaussian noise, otherwise, we invert $\mathbf{P}$ into an intermediate noise level, and a smaller $s$ leads to editing results closer to $\mathbf{P}$. By default, the total number of DDIM steps is set to 64 (we use the same number of sampling steps for our method).

\begin{figure}[htbp]
  \centering
  \includegraphics[width=\linewidth]{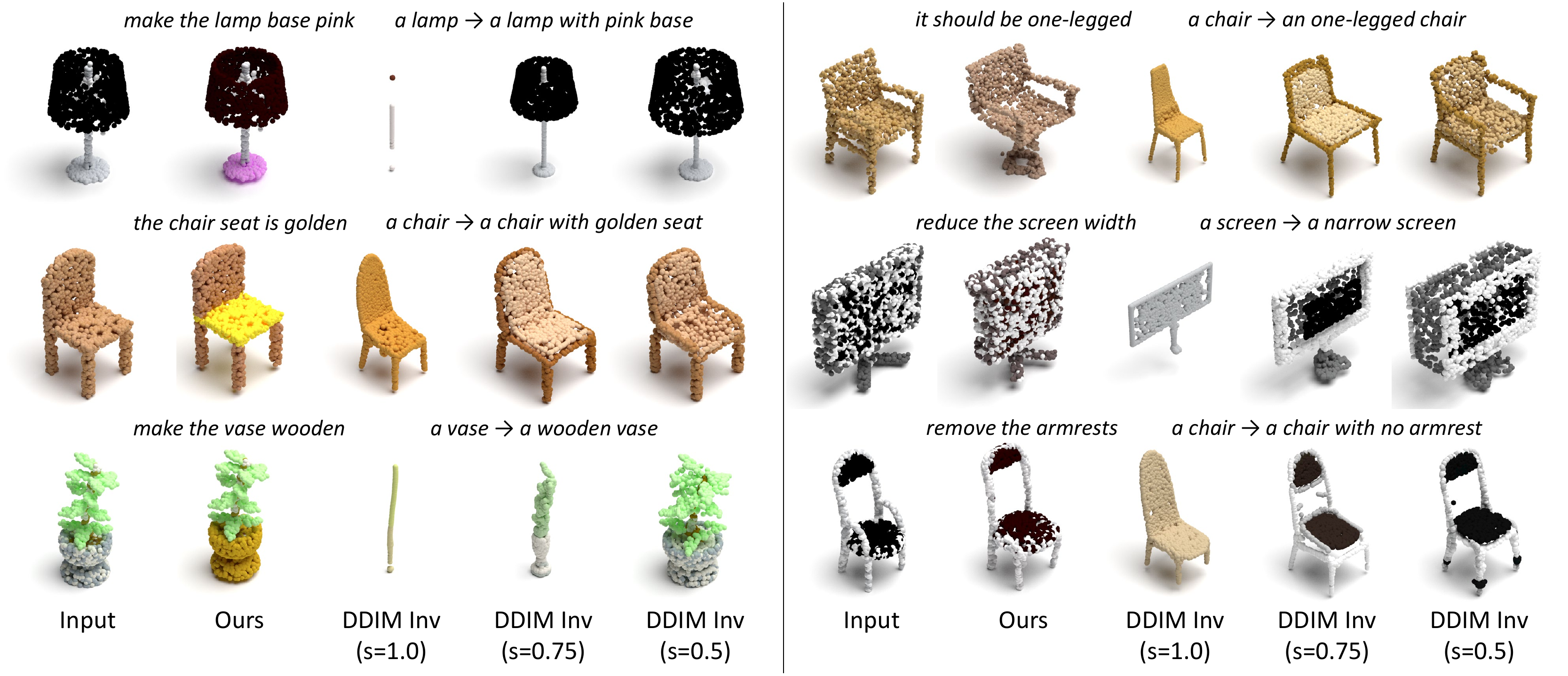}
  \caption{Left: color editing results. Right: geometry editing results. We use the DDIM inversion baseline with 3 different strengths ($s=1.0, 0.75, 0.5$) for comparison. For our method, we annotate the edit instruction above each example. For DDIM inversion baselines, we annotate the used prompts in the format of \emph{``source prompt''->``target prompt''}.}.
  \vspace{-10pt}
  \label{fig:comparison}
\end{figure}

\paragraph{Qualitative Comparison.}
We show the color and geometry editing results in Figure~\ref{fig:comparison}. We can observe that our method shows higher editing quality and accuracy. 
For the DDIM inversion baseline, using a larger strength value for editing can destroy the original shapes. 
And a smaller strength makes no edit to the shapes in most cases. 
Sometimes, the baseline can edit the shape properly (such as the ``remove armrests'' example in the \nth{3} row, the last column). However, our method still shows a better ability to preserve the shape details irrelevant to the desired edit (\nth{3} row, \nth{7} column).

\begin{figure}[htbp]
  \centering
  \vspace{-10pt}
  \includegraphics[width=\linewidth]{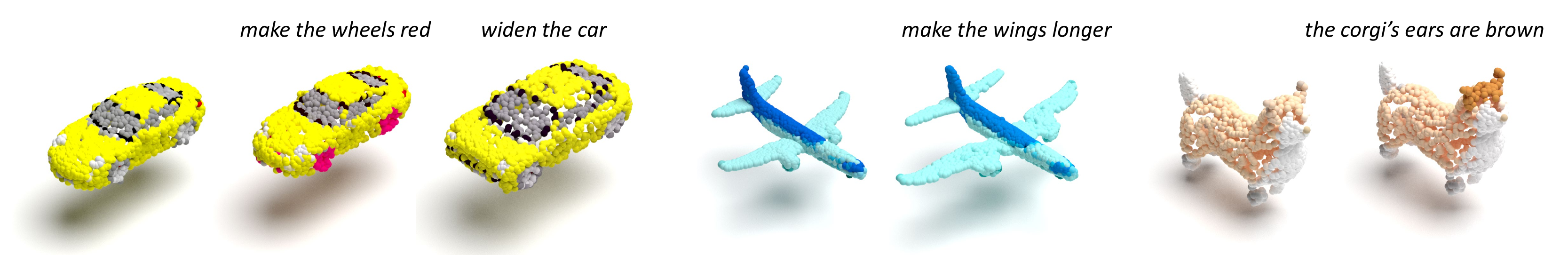}
  \caption{The generalization ability of our InstructP2P model. We show 3 examples from the ``car'', ``airplane'', and ``corgi'' categories. Although these 3 categories are \textbf{not} in our training dataset, our model still generates proper color and geometry manipulations following the edit instructions.}
  \vspace{-10pt}
  \label{fig:generalization}
\end{figure}

\paragraph{Generalization Ability.}
Our InstructP2P is fine-tuned from the Point-E model, which was originally trained on millions of 3D models. Thanks to the generation ability and rich internal 3D priors of Point-E, our shape editing model also exhibits a degree of generalization ability to unseen shape categories and instructions, even though the scale and diversity of the training dataset are limited. As Figure~\ref{fig:generalization} shows, our model can successfully edit the car, the airplane, and the corgi, although these three shape categories are \textbf{not} in our training dataset. The results demonstrate the generalization ability of our method. To the best of our knowledge, our InstructP2P is the first learning-based 3D shape editing method that shows generalization ability across unseen shape categories. We believe our method can serve as a starting point for open-vocabulary 3D shape editing.

\paragraph{Sequential Editing.}
Our InstructP2P also enables sequential point cloud editing by executing a sequence of edit instructions step-by-step. We show an example in Figure~\ref{fig:serialized}, where we gradually edit a short chair without armrests into a tall chair with sky-blue armrests and golden legs. At each step, we edit the color or geometry of a specific chair part, while faithfully preserving the geometric structure and visual appearance of other regions. We believe such a serialized editing ability can be very practical in real-world applications.


\subsection{Quantitative Evaluation}
\label{sec:exp_quan}

\begin{table}[htbp]
\footnotesize
\caption{Quantitative comparison with DDIM inversion. We report the Chamfer-$L_1$ distance for the geometry editing task in the first row, and the RGB mean squared error for the color editing task.}
\vspace{-2pt}
\label{tab:results}
\centering
\begin{tabular}{lcccc}
\toprule
\multirow{2}{*}{Metric} & \multicolumn{3}{c}{DDIM Inv} & \multirow{2}{*}{Ours} \\
\cmidrule(r){2-4}
& $s=1.0$ & $s=0.75$ & $s=0.5$ & \\
\midrule
Chamfer-$L_1$ ($\times 10^{-2}$) $\downarrow$ & 9.72 & 6.12 & 10.16 & 4.09 \\
RGB MSE ($\times 10^{-2}$) $\downarrow$ & 59.73 & 99.71 & 89.75 & 3.28\\
\bottomrule
\vspace{-15pt}
\end{tabular}
\end{table}

\begin{figure}[htbp]
  \centering
  \includegraphics[width=\linewidth]{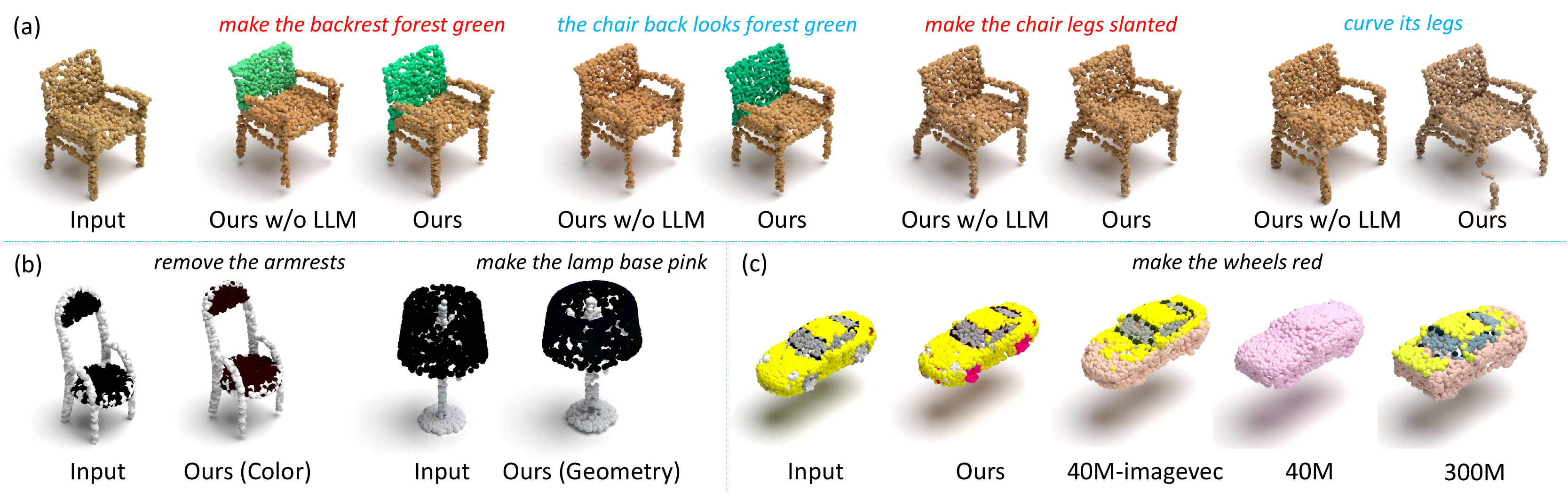}
  \vspace{-5pt}
  \caption{Ablation studies. (a) Ablation on the usage of ChatGPT. We denote the model trained solely with template instructions as ``Ours w/o LLM''. We mark the template instructions in red and the free-style instructions in blue, respectively. (b) Ablation on joint color-geometry training. The models trained on the color and the geometry editing datasets are denoted as ``Ours (Color)'' and ``Ours (Geometry)'' respectively. (c) Ablation on different Point-E variants for fine-tuning.}
  \vspace{-10pt}
  \label{fig:ablation}
\end{figure}

\paragraph{Metrics.} To evaluate our method and the baseline quantitatively, we manually create a color editing test set and a geometry editing test set using the same procedure as in Section~\ref{sec:data}, each containing 100 \emph{<source point cloud, target point cloud, edit instruction>} triplets. To run the DDIM inversion baseline, we also write the corresponding \emph{<source prompt, target prompt>} pairs. 
Although there should be no particular ``ground truth shape'' to judge the editing results, we use the target point cloud as a common reference and consider that a smaller distance to the reference indicates a better editing. 
We leverage two metrics: (i) For geometry editing, we use Chamfer-$L_1$ distance (evaluated with 2048 points) between the $xyz$-coordinates of the generated point cloud and target point cloud as the metric. (ii) For color editing, we compute the mean squared error between the RGB values of the generated point cloud and the target point cloud (we match the two point clouds with the procedure in Section~\ref{sec:training} before computing the error).

\paragraph{Quantitative Comparison.} As shown in Table~\ref{tab:results}, our method achieves a significantly lower Chamfer-$L_1$ distance and RGB error than the DDIM inversion baseline, demonstrating the effectiveness of our method on both the color and geometry editing tasks. For the DDIM inversion baseline, it obtains the worst Chamfer-$L_1$ distance with the strength $s=0.5$ since it tends to ``reconstruct'' the input point cloud without editing. We can also observe the RGB error of our method is much smaller than the DDIM inversion baseline. As Figure~\ref{fig:comparison} has shown, the baseline struggles to edit the color correctly, while our method can edit the colors precisely while preserving the colors of other parts.

\subsection{Ablation Studies}
\label{sec:exp_ablation}

\paragraph{Usage of ChatGPT for Instruction Diversifying.} To demonstrate the efficacy of diversifying edit instructions with ChatGPT, we train an InstructP2P model with only the instructions generated automatically from the template \emph{``make the \{\} \{\}''} (denoted as ``Ours w/o LLM'') for comparison. As Figure~\ref{fig:ablation} (a) shows, although the model trained solely with the template instructions can deal with instructions conforming to the template, it shows a weaker generalization ability to more free-style instructions. In comparison, our model trained with the instructions produced by ChatGPT can understand more diverse edit instructions and generate proper manipulations more consistently.

\paragraph{Joint Color \& Geometry Training.} Our InstructP2P model is trained on the color and geometry editing datasets to handle the two types of editing simultaneously. We then evaluate the model's editing ability when we train it on the two datasets separately. As Figure~\ref{fig:ablation} (b) shows, the model trained solely with color editing examples fails to edit the geometry, and vice versa. By training the model on the two datasets jointly, we can edit both the geometric structure and the visual appearance of the input point cloud with a single model, which makes it more practical.


\paragraph{Varied Point-E Architectures.} We train InstructP2P by fine-tuning the \emph{40M-terxtvec} variant of Point-E, since it is the only variant using languages as conditions. We also tried training the model using different variants for comparison. Although we find that all the models can fit the editing examples in the dataset well, the models fine-tuned from other Point-E variants shows weaker generalization ability to unseen shapes, as shown in Figure~\ref{fig:ablation} (c). We argue that the other Point-E variants take CLIP image embeddings as the condition while we take CLIP text embeddings of edit instructions as the condition during fine-tuning, thus suffering from the gap between the text and image modalities of CLIP embedding space. As a result, the model is overfitted to the training dataset.

\section{Conclusion}

In this paper, we present InstructP2P, the first 3D shape editing framework that can follow human instructions. By generating a collection of shape editing examples and diverse edit instructions with the help of an LLM, we fine-tune a large-scale text-conditioned point cloud diffusion model, \emph{i.e.}, Point-E, into a feed-forward instruction-guided shape editor. The experimental results demonstrate that our InstructP2P is capable of handling both color and geometry editing. Additionally, we also show that our model exhibits generalization abilities to unseen shape categories and instructions, although it is trained on a limited scale of the shape editing dataset.

\paragraph{Limitations.} Our method attains notable point cloud editing results but faces limitations. First, it cannot synthesize complex textures due to point cloud representation constraints. Second, the architecture demands equal input and output sizes, hindering more intuitive geometric operations such as adding/deleting. Finally, fine-grained edits, such as \emph{``increase the height by 0.2m"}, are challenging due to dataset scale and Point-E's limited generation capacity.

\paragraph{Societal Impact.} The emergence of intuitive 3D shape editing techniques has the potential to revolutionize the development of content creation. However, this technique also poses potential risks, such as malicious use in data piracy. Hence, it is crucial to prioritize responsible research and implementation of this technology.

{\small
\bibliographystyle{plainnat}
\bibliography{reference}

\begin{thebibliography}{70}
\providecommand{\natexlab}[1]{#1}
\providecommand{\url}[1]{\texttt{#1}}
\expandafter\ifx\csname urlstyle\endcsname\relax
  \providecommand{\doi}[1]{doi: #1}\else
  \providecommand{\doi}{doi: \begingroup \urlstyle{rm}\Url}\fi

\bibitem[Achlioptas et~al.(2018)Achlioptas, Diamanti, Mitliagkas, and
  Guibas]{achlioptas2018learning}
Panos Achlioptas, Olga Diamanti, Ioannis Mitliagkas, and Leonidas Guibas.
\newblock Learning representations and generative models for 3d point clouds.
\newblock In \emph{International conference on machine learning}, pages 40--49.
  PMLR, 2018.

\bibitem[Achlioptas et~al.(2022)Achlioptas, Huang, Sung, Tulyakov, and
  Guibas]{achlioptas2022changeIt3D}
Panos Achlioptas, Ian Huang, Minhyuk Sung, Sergey Tulyakov, and Leonidas
  Guibas.
\newblock {ChangeIt3D}: Language-assisted 3d shape edits and deformations.
\newblock \emph{https://changeit3d.github.io/}, 2022.

\bibitem[Bar-Tal et~al.(2022)Bar-Tal, Ofri-Amar, Fridman, Kasten, and
  Dekel]{bar2022text2live}
Omer Bar-Tal, Dolev Ofri-Amar, Rafail Fridman, Yoni Kasten, and Tali Dekel.
\newblock Text2live: Text-driven layered image and video editing.
\newblock In \emph{Computer Vision--ECCV 2022: 17th European Conference, Tel
  Aviv, Israel, October 23--27, 2022, Proceedings, Part XV}, pages 707--723.
  Springer, 2022.

\bibitem[Besl and McKay(1992)]{121791}
P.J. Besl and Neil~D. McKay.
\newblock A method for registration of 3-d shapes.
\newblock \emph{IEEE Transactions on Pattern Analysis and Machine
  Intelligence}, 14\penalty0 (2):\penalty0 239--256, 1992.
\newblock \doi{10.1109/34.121791}.

\bibitem[Botsch and Sorkine(2007)]{botsch2007linear}
Mario Botsch and Olga Sorkine.
\newblock On linear variational surface deformation methods.
\newblock \emph{IEEE transactions on visualization and computer graphics},
  14\penalty0 (1):\penalty0 213--230, 2007.

\bibitem[Brooks et~al.(2023)Brooks, Holynski, and
  Efros]{brooks2022instructpix2pix}
Tim Brooks, Aleksander Holynski, and Alexei~A. Efros.
\newblock Instructpix2pix: Learning to follow image editing instructions.
\newblock In \emph{CVPR}, 2023.

\bibitem[Brown et~al.(2020)Brown, Mann, Ryder, Subbiah, Kaplan, Dhariwal,
  Neelakantan, Shyam, Sastry, Askell, et~al.]{brown2020language}
Tom Brown, Benjamin Mann, Nick Ryder, Melanie Subbiah, Jared~D Kaplan, Prafulla
  Dhariwal, Arvind Neelakantan, Pranav Shyam, Girish Sastry, Amanda Askell,
  et~al.
\newblock Language models are few-shot learners.
\newblock \emph{Advances in neural information processing systems},
  33:\penalty0 1877--1901, 2020.

\bibitem[Catmull and Clark(1978)]{catmull1978recursively}
Edwin Catmull and James Clark.
\newblock Recursively generated b-spline surfaces on arbitrary topological
  meshes.
\newblock \emph{Computer-aided design}, 10\penalty0 (6):\penalty0 350--355,
  1978.

\bibitem[Chen et~al.(2022)Chen, Chen, Lei, Zhang, and Jia]{chen2022tango}
Yongwei Chen, Rui Chen, Jiabao Lei, Yabin Zhang, and Kui Jia.
\newblock Tango: Text-driven photorealistic and robust 3d stylization via
  lighting decomposition.
\newblock In \emph{Advances in Neural Information Processing Systems
  (NeurIPS)}, 2022.

\bibitem[Cheng et~al.(2022)Cheng, Lee, Tuyakov, Schwing, and
  Gui]{cheng2022sdfusion}
Yen-Chi Cheng, Hsin-Ying Lee, Sergey Tuyakov, Alex Schwing, and Liangyan Gui.
\newblock {SDFusion}: Multimodal 3d shape completion, reconstruction, and
  generation.
\newblock \emph{arXiv}, 2022.

\bibitem[Chowdhery et~al.(2022)Chowdhery, Narang, Devlin, Bosma, Mishra,
  Roberts, Barham, Chung, Sutton, Gehrmann, et~al.]{chowdhery2022palm}
Aakanksha Chowdhery, Sharan Narang, Jacob Devlin, Maarten Bosma, Gaurav Mishra,
  Adam Roberts, Paul Barham, Hyung~Won Chung, Charles Sutton, Sebastian
  Gehrmann, et~al.
\newblock Palm: Scaling language modeling with pathways.
\newblock \emph{arXiv preprint arXiv:2204.02311}, 2022.

\bibitem[Community(2018)]{blender}
Blender~Online Community.
\newblock \emph{Blender - a 3D modelling and rendering package}.
\newblock Blender Foundation, Stichting Blender Foundation, Amsterdam, 2018.
\newblock URL \url{http://www.blender.org}.

\bibitem[Couairon et~al.(2023)Couairon, Verbeek, Schwenk, and
  Cord]{couairon2023diffedit}
Guillaume Couairon, Jakob Verbeek, Holger Schwenk, and Matthieu Cord.
\newblock Diffedit: Diffusion-based semantic image editing with mask guidance.
\newblock In \emph{The Eleventh International Conference on Learning
  Representations}, 2023.
\newblock URL \url{https://openreview.net/forum?id=3lge0p5o-M-}.

\bibitem[Crowson et~al.(2022)Crowson, Biderman, Kornis, Stander, Hallahan,
  Castricato, and Raff]{crowson2022vqgan}
Katherine Crowson, Stella Biderman, Daniel Kornis, Dashiell Stander, Eric
  Hallahan, Louis Castricato, and Edward Raff.
\newblock Vqgan-clip: Open domain image generation and editing with natural
  language guidance.
\newblock In \emph{Computer Vision--ECCV 2022: 17th European Conference, Tel
  Aviv, Israel, October 23--27, 2022, Proceedings, Part XXXVII}, pages 88--105.
  Springer, 2022.

\bibitem[Devlin et~al.(2019)Devlin, Chang, Lee, and
  Toutanova]{DBLP:conf/naacl/DevlinCLT19}
Jacob Devlin, Ming{-}Wei Chang, Kenton Lee, and Kristina Toutanova.
\newblock {BERT:} pre-training of deep bidirectional transformers for language
  understanding.
\newblock In Jill Burstein, Christy Doran, and Thamar Solorio, editors,
  \emph{Proceedings of the 2019 Conference of the North American Chapter of the
  Association for Computational Linguistics: Human Language Technologies,
  {NAACL-HLT} 2019, Minneapolis, MN, USA, June 2-7, 2019, Volume 1 (Long and
  Short Papers)}, pages 4171--4186. Association for Computational Linguistics,
  2019.
\newblock \doi{10.18653/v1/n19-1423}.
\newblock URL \url{https://doi.org/10.18653/v1/n19-1423}.

\bibitem[Erkoç et~al.(2023)Erkoç, Ma, Shan, Nießner, and
  Dai]{erkoç2023hyperdiffusion}
Ziya Erkoç, Fangchang Ma, Qi~Shan, Matthias Nießner, and Angela Dai.
\newblock Hyperdiffusion: Generating implicit neural fields with weight-space
  diffusion, 2023.

\bibitem[Gao et~al.(2023)Gao, Aigerman, Thibault, Kim, and
  Hanocka]{Gao_2023_SIGGRAPH}
William Gao, Noam Aigerman, Groueix Thibault, Vladimir Kim, and Rana Hanocka.
\newblock Textdeformer: Geometry manipulation using text guidance.
\newblock In \emph{ACM Transactions on Graphics (SIGGRAPH)}, 2023.

\bibitem[Goodfellow et~al.(2020)Goodfellow, Pouget-Abadie, Mirza, Xu,
  Warde-Farley, Ozair, Courville, and Bengio]{goodfellow2020generative}
Ian Goodfellow, Jean Pouget-Abadie, Mehdi Mirza, Bing Xu, David Warde-Farley,
  Sherjil Ozair, Aaron Courville, and Yoshua Bengio.
\newblock Generative adversarial networks.
\newblock \emph{Communications of the ACM}, 63\penalty0 (11):\penalty0
  139--144, 2020.

\bibitem[Gupta et~al.(2023)Gupta, Xiong, Nie, Jones, and
  O{\u{g}}uz]{gupta20233dgen}
Anchit Gupta, Wenhan Xiong, Yixin Nie, Ian Jones, and Barlas O{\u{g}}uz.
\newblock 3dgen: Triplane latent diffusion for textured mesh generation.
\newblock \emph{arXiv preprint arXiv:2303.05371}, 2023.

\bibitem[Hao et~al.(2020)Hao, Averbuch-Elor, Snavely, and
  Belongie]{hao2020dualsdf}
Zekun Hao, Hadar Averbuch-Elor, Noah Snavely, and Serge Belongie.
\newblock Dualsdf: Semantic shape manipulation using a two-level
  representation.
\newblock In \emph{Proceedings of the IEEE/CVF Conference on Computer Vision
  and Pattern Recognition}, pages 7631--7641, 2020.

\bibitem[Haque et~al.(2023)Haque, Tancik, Efros, Holynski, and
  Kanazawa]{haque2023instruct}
Ayaan Haque, Matthew Tancik, Alexei~A Efros, Aleksander Holynski, and Angjoo
  Kanazawa.
\newblock Instruct-nerf2nerf: Editing 3d scenes with instructions.
\newblock \emph{arXiv preprint arXiv:2303.12789}, 2023.

\bibitem[Hertz et~al.(2022)Hertz, Mokady, Tenenbaum, Aberman, Pritch, and
  Cohen-Or]{hertz2022prompt}
Amir Hertz, Ron Mokady, Jay Tenenbaum, Kfir Aberman, Yael Pritch, and Daniel
  Cohen-Or.
\newblock Prompt-to-prompt image editing with cross attention control.
\newblock 2022.

\bibitem[Ho et~al.(2020)Ho, Jain, and Abbeel]{ho2020denoising}
Jonathan Ho, Ajay Jain, and Pieter Abbeel.
\newblock Denoising diffusion probabilistic models.
\newblock \emph{Advances in Neural Information Processing Systems},
  33:\penalty0 6840--6851, 2020.

\bibitem[Hu et~al.(2023)Hu, Hui, Liu, Li, and Fu]{hu2023neural}
Jingyu Hu, Ka-Hei Hui, Zhengzhe Liu, Ruihui Li, and Chi-Wing Fu.
\newblock Neural wavelet-domain diffusion for 3d shape generation, inversion,
  and manipulation.
\newblock \emph{arXiv preprint arXiv:2302.00190}, 2023.

\bibitem[Joshi et~al.(2007)Joshi, Meyer, DeRose, Green, and
  Sanocki]{joshi2007harmonic}
Pushkar Joshi, Mark Meyer, Tony DeRose, Brian Green, and Tom Sanocki.
\newblock Harmonic coordinates for character articulation.
\newblock \emph{ACM transactions on graphics (TOG)}, 26\penalty0 (3):\penalty0
  71--es, 2007.

\bibitem[Ju et~al.(2005)Ju, Schaefer, and Warren]{10.1145/1073204.1073229}
Tao Ju, Scott Schaefer, and Joe Warren.
\newblock Mean value coordinates for closed triangular meshes.
\newblock \emph{ACM Trans. Graph.}, 24\penalty0 (3):\penalty0 561–566, jul
  2005.
\newblock ISSN 0730-0301.
\newblock \doi{10.1145/1073204.1073229}.
\newblock URL \url{https://doi.org/10.1145/1073204.1073229}.

\bibitem[Kamata et~al.(2023)Kamata, Sakuma, Hayakawa, Ishii, and
  Narihira]{kamata2023instruct}
Hiromichi Kamata, Yuiko Sakuma, Akio Hayakawa, Masato Ishii, and Takuya
  Narihira.
\newblock Instruct 3d-to-3d: Text instruction guided 3d-to-3d conversion.
\newblock \emph{arXiv preprint arXiv:2303.15780}, 2023.

\bibitem[Karras et~al.(2020)Karras, Laine, Aittala, Hellsten, Lehtinen, and
  Aila]{Karras_2020_CVPR}
Tero Karras, Samuli Laine, Miika Aittala, Janne Hellsten, Jaakko Lehtinen, and
  Timo Aila.
\newblock Analyzing and improving the image quality of stylegan.
\newblock In \emph{Proceedings of the IEEE/CVF Conference on Computer Vision
  and Pattern Recognition (CVPR)}, June 2020.

\bibitem[Kawar et~al.(2023)Kawar, Zada, Lang, Tov, Chang, Dekel, Mosseri, and
  Irani]{kawar2023imagic}
Bahjat Kawar, Shiran Zada, Oran Lang, Omer Tov, Huiwen Chang, Tali Dekel, Inbar
  Mosseri, and Michal Irani.
\newblock Imagic: Text-based real image editing with diffusion models.
\newblock In \emph{Conference on Computer Vision and Pattern Recognition 2023},
  2023.

\bibitem[Kim et~al.(2022)Kim, Kwon, and Ye]{Kim_2022_CVPR}
Gwanghyun Kim, Taesung Kwon, and Jong~Chul Ye.
\newblock Diffusionclip: Text-guided diffusion models for robust image
  manipulation.
\newblock In \emph{Proceedings of the IEEE/CVF Conference on Computer Vision
  and Pattern Recognition (CVPR)}, pages 2426--2435, June 2022.

\bibitem[Lee et~al.(2023)Lee, Jang, Chen, Qiu, and Huang]{lee2023shape}
Yao-Chih Lee, Ji-Ze~Genevieve Jang, Yi-Ting Chen, Elizabeth Qiu, and Jia-Bin
  Huang.
\newblock Shape-aware text-driven layered video editing.
\newblock \emph{arXiv e-prints}, pages arXiv--2301, 2023.

\bibitem[Li et~al.(2022)Li, Duan, Zhou, and Lu]{li2022diffusion}
Muheng Li, Yueqi Duan, Jie Zhou, and Jiwen Lu.
\newblock Diffusion-sdf: Text-to-shape via voxelized diffusion.
\newblock \emph{arXiv preprint arXiv:2212.03293}, 2022.

\bibitem[Li et~al.(2023)Li, Dou, Chen, Ni, Sun, Liu, and Wang]{li20233dqd}
Yuhan Li, Yishun Dou, Xuanhong Chen, Bingbing Ni, Yilin Sun, Yutian Liu, and
  Fuzhen Wang.
\newblock 3dqd: Generalized deep 3d shape prior via part-discretized diffusion
  process.
\newblock \emph{arXiv preprint arXiv:2303.10406}, 2023.

\bibitem[Liu et~al.(2021)Liu, Sung, Mech, and Su]{liu2021deepmetahandles}
Minghua Liu, Minhyuk Sung, Radomir Mech, and Hao Su.
\newblock Deepmetahandles: Learning deformation meta-handles of 3d meshes with
  biharmonic coordinates.
\newblock In \emph{Proceedings of the IEEE/CVF Conference on Computer Vision
  and Pattern Recognition}, pages 12--21, 2021.

\bibitem[Liu et~al.(2023)Liu, Feng, Black, Nowrouzezahrai, Paull, and
  Liu]{Liu2023MeshDiffusion}
Zhen Liu, Yao Feng, Michael~J. Black, Derek Nowrouzezahrai, Liam Paull, and
  Weiyang Liu.
\newblock Meshdiffusion: Score-based generative 3d mesh modeling.
\newblock In \emph{International Conference on Learning Representations}, 2023.
\newblock URL \url{https://openreview.net/forum?id=0cpM2ApF9p6}.

\bibitem[Loop(1987)]{Loop1987}
Charles Loop.
\newblock Smooth subdivision surfaces based on triangles.
\newblock Master's thesis, Department of Mathematics, University of Utah, 1987.

\bibitem[Luo and Hu(2021)]{Luo_2021_CVPR}
Shitong Luo and Wei Hu.
\newblock Diffusion probabilistic models for 3d point cloud generation.
\newblock In \emph{Proceedings of the IEEE/CVF Conference on Computer Vision
  and Pattern Recognition (CVPR)}, pages 2837--2845, June 2021.

\bibitem[Lyu et~al.(2023{\natexlab{a}})Lyu, Lin, Li, He, Dong, and
  Tan]{lyu2023deltaedit}
Yueming Lyu, Tianwei Lin, Fu~Li, Dongliang He, Jing Dong, and Tieniu Tan.
\newblock Deltaedit: Exploring text-free training for text-driven image
  manipulation.
\newblock \emph{arXiv preprint arXiv:2303.06285}, 2023{\natexlab{a}}.

\bibitem[Lyu et~al.(2023{\natexlab{b}})Lyu, Wang, An, Zhang, Lin, and
  Dai]{lyu2023controllable}
Zhaoyang Lyu, Jinyi Wang, Yuwei An, Ya~Zhang, Dahua Lin, and Bo~Dai.
\newblock Controllable mesh generation through sparse latent point diffusion
  models.
\newblock \emph{arXiv preprint arXiv:2303.07938}, 2023{\natexlab{b}}.

\bibitem[Ma et~al.(2023)Ma, Zhang, Sun, Ji, Wang, Jiang, Zhuang, and
  Ji]{ma2023xmesh}
Yiwei Ma, Xiaioqing Zhang, Xiaoshuai Sun, Jiayi Ji, Haowei Wang, Guannan Jiang,
  Weilin Zhuang, and Rongrong Ji.
\newblock X-mesh: Towards fast and accurate text-driven 3d stylization via
  dynamic textual guidance, 2023.

\bibitem[Melas-Kyriazi et~al.(2023)Melas-Kyriazi, Rupprecht, and
  Vedaldi]{melaskyriazi2023projection}
Luke Melas-Kyriazi, Christian Rupprecht, and Andrea Vedaldi.
\newblock Pc2: Projection-conditioned point cloud diffusion for single-image 3d
  reconstruction.
\newblock In \emph{Arxiv}, 2023.

\bibitem[Meng et~al.(2022)Meng, He, Song, Song, Wu, Zhu, and
  Ermon]{meng2022sdedit}
Chenlin Meng, Yutong He, Yang Song, Jiaming Song, Jiajun Wu, Jun-Yan Zhu, and
  Stefano Ermon.
\newblock {SDE}dit: Guided image synthesis and editing with stochastic
  differential equations.
\newblock In \emph{International Conference on Learning Representations}, 2022.
\newblock URL \url{https://openreview.net/forum?id=aBsCjcPu_tE}.

\bibitem[Mescheder et~al.(2019)Mescheder, Oechsle, Niemeyer, Nowozin, and
  Geiger]{mescheder2019occupancy}
Lars Mescheder, Michael Oechsle, Michael Niemeyer, Sebastian Nowozin, and
  Andreas Geiger.
\newblock Occupancy networks: Learning 3d reconstruction in function space.
\newblock In \emph{Proceedings of the IEEE/CVF conference on computer vision
  and pattern recognition}, pages 4460--4470, 2019.

\bibitem[Michel et~al.(2022)Michel, Bar-On, Liu, Benaim, and
  Hanocka]{Michel_2022_CVPR}
Oscar Michel, Roi Bar-On, Richard Liu, Sagie Benaim, and Rana Hanocka.
\newblock Text2mesh: Text-driven neural stylization for meshes.
\newblock In \emph{Proceedings of the IEEE/CVF Conference on Computer Vision
  and Pattern Recognition (CVPR)}, pages 13492--13502, June 2022.

\bibitem[Mikolov et~al.(2013)Mikolov, Sutskever, Chen, Corrado, and
  Dean]{mikolov2013distributed}
Tomas Mikolov, Ilya Sutskever, Kai Chen, Greg~S Corrado, and Jeff Dean.
\newblock Distributed representations of words and phrases and their
  compositionality.
\newblock \emph{Advances in neural information processing systems}, 26, 2013.

\bibitem[Mo et~al.(2019)Mo, Zhu, Chang, Yi, Tripathi, Guibas, and
  Su]{Mo_2019_CVPR}
Kaichun Mo, Shilin Zhu, Angel~X. Chang, Li~Yi, Subarna Tripathi, Leonidas~J.
  Guibas, and Hao Su.
\newblock Partnet: A large-scale benchmark for fine-grained and hierarchical
  part-level 3d object understanding.
\newblock In \emph{Proceedings of the IEEE/CVF Conference on Computer Vision
  and Pattern Recognition (CVPR)}, June 2019.

\bibitem[Mokady et~al.(2022)Mokady, Hertz, Aberman, Pritch, and
  Cohen-Or]{mokady2022null}
Ron Mokady, Amir Hertz, Kfir Aberman, Yael Pritch, and Daniel Cohen-Or.
\newblock Null-text inversion for editing real images using guided diffusion
  models.
\newblock \emph{arXiv preprint arXiv:2211.09794}, 2022.

\bibitem[Nakayama et~al.(2023)Nakayama, Uy, Huang, Hu, Li, and
  Guibas]{nakayama2023difffacto}
Kiyohiro Nakayama, Mikaela~Angelina Uy, Jiahui Huang, Shi-Min Hu, Ke~Li, and
  Leonidas~J Guibas.
\newblock Difffacto controllable part-based 3d point cloud generation with
  cross diffusion.
\newblock \emph{arXiv preprint arXiv:2305.01921}, 2023.

\bibitem[Nichol et~al.(2022)Nichol, Jun, Dhariwal, Mishkin, and
  Chen]{nichol2022point}
Alex Nichol, Heewoo Jun, Prafulla Dhariwal, Pamela Mishkin, and Mark Chen.
\newblock Point-e: A system for generating 3d point clouds from complex
  prompts.
\newblock \emph{arXiv preprint arXiv:2212.08751}, 2022.

\bibitem[OpenAI(2022)]{OpenAI2022ChatGPT}
OpenAI.
\newblock Introducing chatgpt, 2022.
\newblock URL \url{https://openai.com/blog/chatgpt}.

\bibitem[OpenAI(2023)]{openai2023gpt4}
OpenAI.
\newblock Gpt-4 technical report, 2023.

\bibitem[Park et~al.(2019)Park, Florence, Straub, Newcombe, and
  Lovegrove]{park2019deepsdf}
Jeong~Joon Park, Peter Florence, Julian Straub, Richard Newcombe, and Steven
  Lovegrove.
\newblock Deepsdf: Learning continuous signed distance functions for shape
  representation.
\newblock In \emph{Proceedings of the IEEE/CVF conference on computer vision
  and pattern recognition}, pages 165--174, 2019.

\bibitem[Patashnik et~al.(2021)Patashnik, Wu, Shechtman, Cohen-Or, and
  Lischinski]{patashnik2021styleclip}
Or~Patashnik, Zongze Wu, Eli Shechtman, Daniel Cohen-Or, and Dani Lischinski.
\newblock Styleclip: Text-driven manipulation of stylegan imagery.
\newblock In \emph{Proceedings of the IEEE/CVF International Conference on
  Computer Vision}, pages 2085--2094, 2021.

\bibitem[Pearl et~al.(2022)Pearl, Lang, Hu, Yeh, and Hanocka]{pearl2022geocode}
Ofek Pearl, Itai Lang, Yuhua Hu, Raymond~A Yeh, and Rana Hanocka.
\newblock Geocode: Interpretable shape programs.
\newblock \emph{arXiv preprint arXiv:2212.11715}, 2022.

\bibitem[Pennington et~al.(2014)Pennington, Socher, and
  Manning]{pennington2014glove}
Jeffrey Pennington, Richard Socher, and Christopher~D Manning.
\newblock Glove: Global vectors for word representation.
\newblock In \emph{Proceedings of the 2014 conference on empirical methods in
  natural language processing (EMNLP)}, pages 1532--1543, 2014.

\bibitem[Radford et~al.(2018)Radford, Narasimhan, Salimans, Sutskever,
  et~al.]{radford2018improving}
Alec Radford, Karthik Narasimhan, Tim Salimans, Ilya Sutskever, et~al.
\newblock Improving language understanding by generative pre-training.
\newblock 2018.

\bibitem[Radford et~al.(2021)Radford, Kim, Hallacy, Ramesh, Goh, Agarwal,
  Sastry, Askell, Mishkin, Clark, et~al.]{radford2021learning}
Alec Radford, Jong~Wook Kim, Chris Hallacy, Aditya Ramesh, Gabriel Goh,
  Sandhini Agarwal, Girish Sastry, Amanda Askell, Pamela Mishkin, Jack Clark,
  et~al.
\newblock Learning transferable visual models from natural language
  supervision.
\newblock In \emph{International Conference on Machine Learning}, pages
  8748--8763. PMLR, 2021.

\bibitem[Rombach et~al.(2022)Rombach, Blattmann, Lorenz, Esser, and
  Ommer]{Rombach_2022_CVPR}
Robin Rombach, Andreas Blattmann, Dominik Lorenz, Patrick Esser, and Bj\"orn
  Ommer.
\newblock High-resolution image synthesis with latent diffusion models.
\newblock In \emph{Proceedings of the IEEE/CVF Conference on Computer Vision
  and Pattern Recognition (CVPR)}, pages 10684--10695, June 2022.

\bibitem[Saharia et~al.(2022)Saharia, Chan, Saxena, Li, Whang, Denton,
  Ghasemipour, Ayan, Mahdavi, Lopes, et~al.]{saharia2022photorealistic}
Chitwan Saharia, William Chan, Saurabh Saxena, Lala Li, Jay Whang, Emily
  Denton, Seyed Kamyar~Seyed Ghasemipour, Burcu~Karagol Ayan, S~Sara Mahdavi,
  Rapha~Gontijo Lopes, et~al.
\newblock Photorealistic text-to-image diffusion models with deep language
  understanding.
\newblock \emph{arXiv preprint arXiv:2205.11487}, 2022.

\bibitem[Shue et~al.(2022)Shue, Chan, Po, Ankner, Wu, and
  Wetzstein]{shue20223d}
J~Ryan Shue, Eric~Ryan Chan, Ryan Po, Zachary Ankner, Jiajun Wu, and Gordon
  Wetzstein.
\newblock 3d neural field generation using triplane diffusion.
\newblock \emph{arXiv preprint arXiv:2211.16677}, 2022.

\bibitem[Song et~al.(2021)Song, Meng, and Ermon]{song2021denoising}
Jiaming Song, Chenlin Meng, and Stefano Ermon.
\newblock Denoising diffusion implicit models.
\newblock In \emph{International Conference on Learning Representations}, 2021.
\newblock URL \url{https://openreview.net/forum?id=St1giarCHLP}.

\bibitem[Sorkine et~al.(2004)Sorkine, Cohen-Or, Lipman, Alexa, R{\"o}ssl, and
  Seidel]{sorkine2004laplacian}
Olga Sorkine, Daniel Cohen-Or, Yaron Lipman, Marc Alexa, Christian R{\"o}ssl,
  and H-P Seidel.
\newblock Laplacian surface editing.
\newblock In \emph{Proceedings of the 2004 Eurographics/ACM SIGGRAPH symposium
  on Geometry processing}, pages 175--184, 2004.

\bibitem[Tang et~al.(2022)Tang, Lev, Bi, Justus, and
  Nie{\ss}ner]{tang2022neural}
Jiapeng Tang, Markhasin Lev, Wang Bi, Thies Justus, and Matthias Nie{\ss}ner.
\newblock Neural shape deformation priors.
\newblock In \emph{Advances in Neural Information Processing Systems}, 2022.

\bibitem[Tyszkiewicz et~al.(2023)Tyszkiewicz, Fua, and
  Trulls]{tyszkiewicz2023gecco}
Micha{\l}~J Tyszkiewicz, Pascal Fua, and Eduard Trulls.
\newblock Gecco: Geometrically-conditioned point diffusion models.
\newblock \emph{arXiv preprint arXiv:2303.05916}, 2023.

\bibitem[Virtanen et~al.(2020)Virtanen, Gommers, Oliphant, Haberland, Reddy,
  Cournapeau, Burovski, Peterson, Weckesser, Bright, {van der Walt}, Brett,
  Wilson, Millman, Mayorov, Nelson, Jones, Kern, Larson, Carey, Polat, Feng,
  Moore, {VanderPlas}, Laxalde, Perktold, Cimrman, Henriksen, Quintero, Harris,
  Archibald, Ribeiro, Pedregosa, {van Mulbregt}, and {SciPy 1.0
  Contributors}]{2020SciPy-NMeth}
Pauli Virtanen, Ralf Gommers, Travis~E. Oliphant, Matt Haberland, Tyler Reddy,
  David Cournapeau, Evgeni Burovski, Pearu Peterson, Warren Weckesser, Jonathan
  Bright, St{\'e}fan~J. {van der Walt}, Matthew Brett, Joshua Wilson, K.~Jarrod
  Millman, Nikolay Mayorov, Andrew R.~J. Nelson, Eric Jones, Robert Kern, Eric
  Larson, C~J Carey, {\.I}lhan Polat, Yu~Feng, Eric~W. Moore, Jake
  {VanderPlas}, Denis Laxalde, Josef Perktold, Robert Cimrman, Ian Henriksen,
  E.~A. Quintero, Charles~R. Harris, Anne~M. Archibald, Ant{\^o}nio~H. Ribeiro,
  Fabian Pedregosa, Paul {van Mulbregt}, and {SciPy 1.0 Contributors}.
\newblock {{SciPy} 1.0: Fundamental Algorithms for Scientific Computing in
  Python}.
\newblock \emph{Nature Methods}, 17:\penalty0 261--272, 2020.
\newblock \doi{10.1038/s41592-019-0686-2}.

\bibitem[Yildirim et~al.(2023)Yildirim, Baday, Erdem, Erdem, and
  Dundar]{yildirim2023inst}
Ahmet~Burak Yildirim, Vedat Baday, Erkut Erdem, Aykut Erdem, and Aysegul
  Dundar.
\newblock Inst-inpaint: Instructing to remove objects with diffusion models.
\newblock \emph{arXiv preprint arXiv:2304.03246}, 2023.

\bibitem[Zeng et~al.(2022)Zeng, Vahdat, Williams, Gojcic, Litany, Fidler, and
  Kreis]{zeng2022lion}
Xiaohui Zeng, Arash Vahdat, Francis Williams, Zan Gojcic, Or~Litany, Sanja
  Fidler, and Karsten Kreis.
\newblock Lion: Latent point diffusion models for 3d shape generation.
\newblock In \emph{Advances in Neural Information Processing Systems
  (NeurIPS)}, 2022.

\bibitem[Zhang et~al.(2022)Zhang, Han, Ghosh, Metaxas, and Ren]{zhang2022sine}
Zhixing Zhang, Ligong Han, Arnab Ghosh, Dimitris Metaxas, and Jian Ren.
\newblock Sine: Single image editing with text-to-image diffusion models.
\newblock \emph{arXiv preprint arXiv:2212.04489}, 2022.

\bibitem[Zheng et~al.(2022)Zheng, Li, Guo, Wan, and Wang]{zheng2022bridging}
Wanfeng Zheng, Qiang Li, Xiaoyan Guo, Pengfei Wan, and Zhongyuan Wang.
\newblock Bridging clip and stylegan through latent alignment for image
  editing.
\newblock \emph{arXiv preprint arXiv:2210.04506}, 2022.

\bibitem[Zhou et~al.(2021)Zhou, Du, and Wu]{Zhou_2021_ICCV}
Linqi Zhou, Yilun Du, and Jiajun Wu.
\newblock 3d shape generation and completion through point-voxel diffusion.
\newblock In \emph{Proceedings of the IEEE/CVF International Conference on
  Computer Vision (ICCV)}, pages 5826--5835, October 2021.

\end{thebibliography}
}


\clearpage
\clearpage
\newpage

\renewcommand\thesection{\Alph{section}}
\setcounter{section}{0}

\section{Details of The Shape Editing Dataset}
\label{sec:supp_dataset}

\subsection{Dataset Generation Algorithms}
To generate the color editing examples, we use the part-level annotations of the PartNet~\cite{Mo_2019_CVPR} dataset. For each point cloud in PartNet, we assign a random color to each part to get the target point clouds. To generate the geometry editing examples, we use the blender shape programs of 3 categories (chair, vase, table) provided by~\citet{pearl2022geocode}~\footnote{\url{https://github.com/threedle/GeoCode}}. We use the shape programs to generate a set of meshes first, then we randomly alter each editable parameter and synthesize a target mesh for each edited parameter with the shape programs. Finally, we sample the source and target point clouds from the source and target meshes. To be noted, we pre-dine a ``parameter inscrease instruction'' and a ``parameter decrease instruction'' for each editable shape parameter to generate the simple edit instructions. We give the detailed dataset generation algorithms for the color and geometry editing datasets in Algorithm~\ref{alg:color} and~\ref{alg:geometry}, respectively. For simplicity, Algorithm~\ref{alg:geometry} only takes a single category into account, but the procedures are equal for all the 3 categories.

\subsection{ChatGPT Prompt}
\begin{figure}[h]
  \centering
  \vspace{-6mm}
  \includegraphics[width=\linewidth]{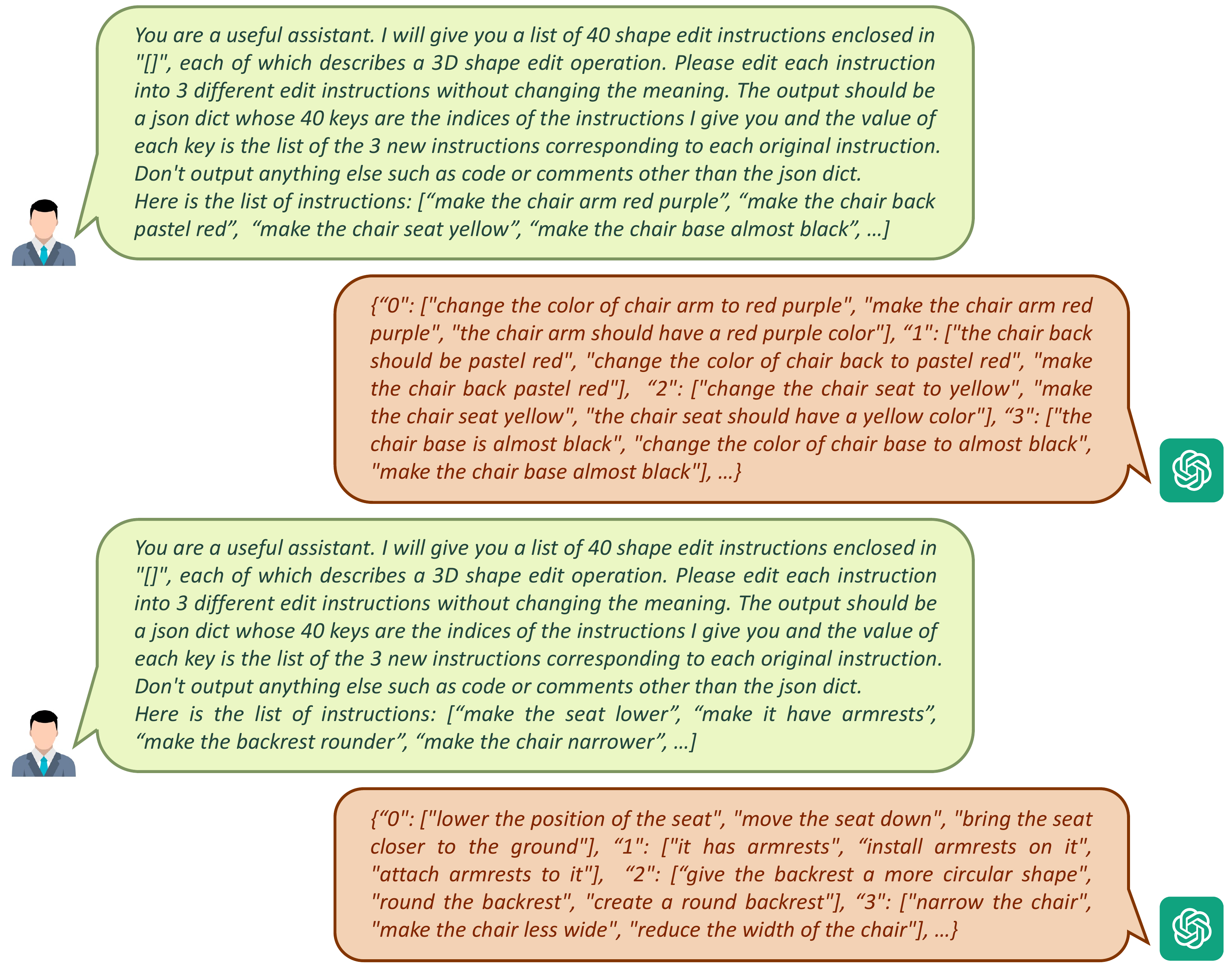}
  \vspace{-6mm}
  \caption{Our prompt for calling the ChatGPT API to generate diverse edit instructions and the responses from ChatGPT. For each API call, we process $40$ edit instructions in parallel to maximize the number of input tokens and reduce cost. The output format is a JSON dict.}
  \label{fig:chatgpt}
\end{figure}

We generate an associated edit instruction for each pair of edit examples using the template \emph{``Make the \textcolor{red}{\{\}} \textcolor{cyan}{\{\}}''}, where \emph{\textcolor{red}{\{\}}} denotes the editing objective and \emph{\textcolor{cyan}{\{\}}} is a description of the editing. To diversify these naive edit instructions, we leverage the ChatGPT~\cite{OpenAI2022ChatGPT} API to rewrite each edit instruction into 3 different instructions sharing the same meaning. We show our prompt for calling the API in Figure~\ref{fig:chatgpt}. To reduce the API cost, we process $40$ edit instructions in parallel at each API call to maximize the number of input tokens, and we request ChatGPT to output a JSON dict whose keys are the indices of the input instructions while the values are the lists of 3 rewritten instructions.

\subsection{Dataset Details}

\begin{figure}[h]
  \centering
  \includegraphics[width=\linewidth]{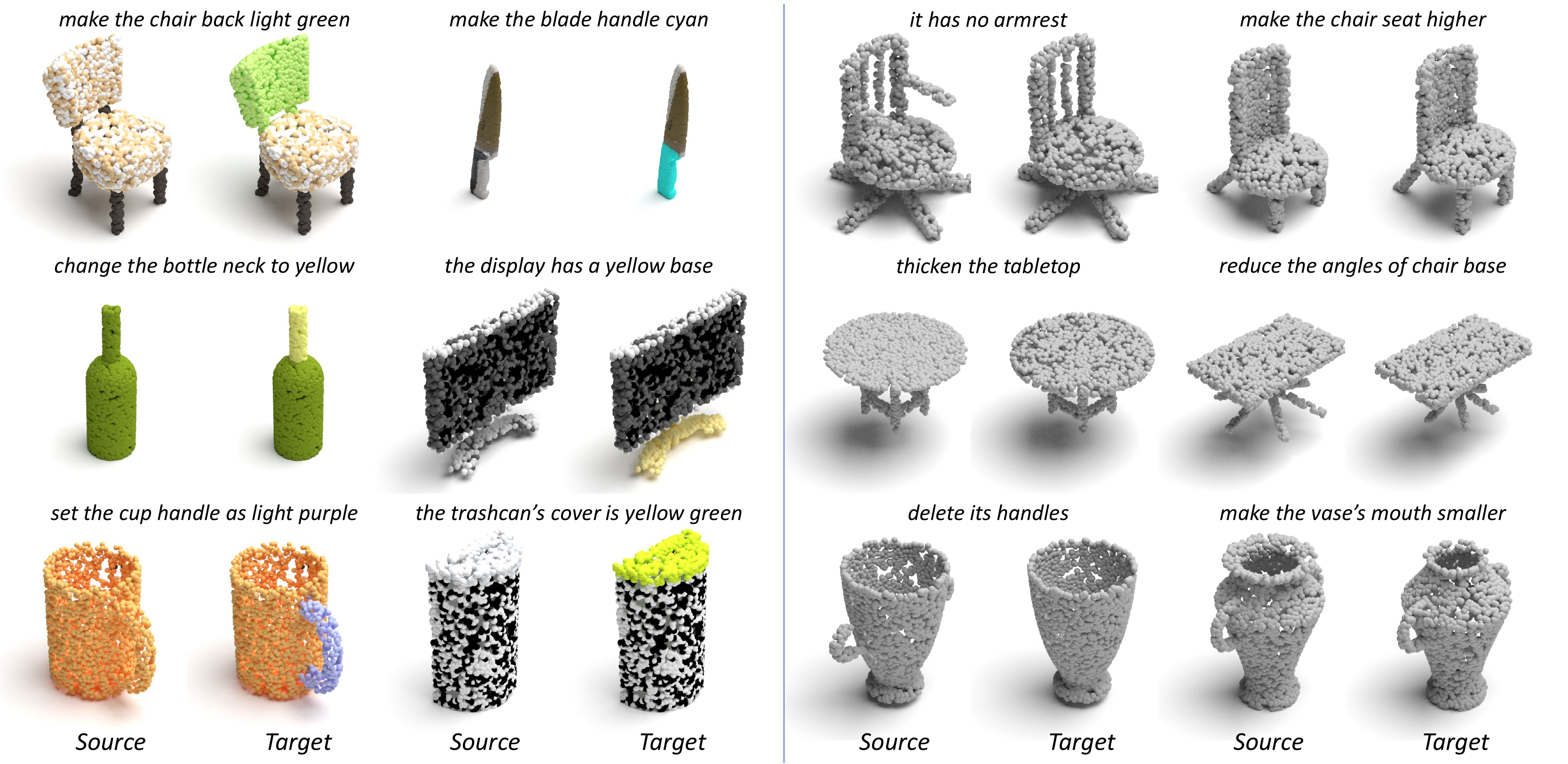}
  \vspace{-6mm}
  \caption{Left: samples from our color editing dataset; Right: samples from our geometry editing dataset.}
  \label{fig:dataset}
\end{figure}

After running Algorithm~\ref{alg:color} and~\ref{alg:geometry} for dataset generation, we obtain 98763 color editing examples and 80270 geometry editing examples in total. For each pair of editing example, we have a source point cloud, a target point cloud, and 3 different edit instructions produced by ChatGPT. For the color editing dataset, we utilize the first-level part annotations of PartNet. All the shape categories and the corresponding part categories we used are listed in Table~\ref{tab:partnet}. For the geometry editing dataset, we list all the editable shape parameters and the shape properties they control for the three categories (chair, vase, table) in Table~\ref{tab:geocode}. Finally, we visualize samples from our instruction-guided point cloud editing dataset in Figure~\ref{fig:dataset}. 

\clearpage
\clearpage
\newpage

\begin{algorithm}[h]
\caption{Generating Color Editing Examples}
\label{alg:color}
\textbf{Input:} Point clouds $\{\mathbf{P}_i\}_{i=1}^M$, part annotations $\{\{\mathbf{l}_{i,j}\}_{j=1}^{N_i}\}_{i=1}^{M}$ where $N_i$ denotes the number of parts of $\mathbf{P}_i$, colors set $C=\{\mathbf{c}_k\}_{k=1}^{N_c}$. \\
\textbf{Output:} Color editing dataset $D_{\text{color}}.$
\begin{algorithmic}[1]
\STATE $D_{\text{color}} \gets []$
\FOR{$i = 1$ \TO $M$}
    \FOR{$j = 1$ \TO $N_i$}
        \STATE $\mathbf{c}_k \gets \textsc{Random}(C)$
        \STATE $\mathbf{P}'_{i,j} \gets \textsc{Assign}(\mathbf{P}_i, \mathbf{l}_{i,j}, \mathbf{c}_k)$ \COMMENT{Assign color $\mathbf{c}_k$ to the part $\mathbf{l}_{i,j}$ of $\mathbf{P}_i$}
        \STATE $n_{\text{part}} \gets \textsc{Name}(\mathbf{L}_{i,j})$
        \STATE $n_{\text{color}} \gets \textsc{Name}(\mathbf{c}_{k})$
        \STATE $text \gets \textsc{Template}(n_{\text{part}}, n_{\text{color}})$ \COMMENT{Generate edit instruction using text template}
        \STATE $\textbf{Add} <\mathbf{P}_i, \mathbf{P}'_{i,j}, text>$ to $D_{\text{color}}$
    \ENDFOR
\ENDFOR
\RETURN $D_{\text{color}}$
\end{algorithmic}
\end{algorithm}

\begin{algorithm}[h]
\caption{Generating Geometry Editing Examples (Single Category)}
\label{alg:geometry}
\textbf{Input:} Meshes $\{\mathbf{S}_i\}_{i=1}^M$, shape parameters $\{\mathbf{e}_{j}\}_{j=1}^{N}$ where $N$ denotes the number of editable parameters of this category, parameter increase instructions $\{\mathbf{t}_{j}^{\text{inc}}\}_{j=1}^{N}$, parameter decrease instructions $\{\mathbf{t}_{j}^{\text{dec}}\}_{j=1}^{N}$, shape program $G$ of this category. \\
\textbf{Output:} Geometry editing dataset $D_{\text{geometry}}.$
\begin{algorithmic}[1]
\STATE $D_{\text{geometry}} \gets []$
\FOR{$i = 1$ \TO $M$}
    \FOR{$j = 1$ \TO $N$}
        \STATE $\mathbf{v}_{i,j} \gets \textsc{Value}(\mathbf{S}_i, \mathbf{e}_j)$ \COMMENT{Get the original parameter value}
        \STATE $\mathbf{v}'_{i,j} \gets \textsc{RandomChange}(\mathbf{v}_{i,j})$ \COMMENT{Get the edited parameter value randomly}
        \STATE $\mathbf{S}'_{i,j} \gets \textsc{Program}(G, \mathbf{S}_i, \mathbf{v}'_{i,j})$ \COMMENT{Generate the edited mesh using the shape program}
        \IF{$\mathbf{v}'_{i,j} > \mathbf{v}_{i,j}$}
            \STATE $text \gets \mathbf{t}_{j}^{\text{inc}}$
        \ELSE
            \STATE $text \gets \mathbf{t}_{j}^{\text{dec}}$
        \ENDIF
        \STATE $\mathbf{P}_{i} \gets \textsc{Sample}(\mathbf{S}_{i})$ \COMMENT{Sample the source point cloud}
        \STATE $\mathbf{P}'_{i,j} \gets \textsc{Sample}(\mathbf{S}'_{i,j})$ \COMMENT{Sample the target point cloud}
        \STATE $\textbf{Add} <\mathbf{P}_i, \mathbf{P}'_{i,j}, text> $ to $D_{\text{geometry}}$
    \ENDFOR
\ENDFOR
\RETURN $D_{\text{geometry}}$
\end{algorithmic}
\end{algorithm}

\clearpage
\clearpage
\newpage

\clearpage
\clearpage
\newpage

\begin{table}[h]
\scriptsize
\renewcommand{\arraystretch}{1.0}
\renewcommand{\tabcolsep}{0mm}
\centering
\caption{The shape categories and the corresponding part categories we used for generating the color editing dataset.}
\begin{tabular}{cccccccccccc}
\toprule
Bag & Bed & Bowl & Bottle & Chair & Clock & Dishwasher & Display & Door & Earphone & Faucet & Hat \\
\midrule
body & sleep area & container & lid & head & base & body & screen & body & earbud unit & switch & crown \\
handle & ladder & bottom & handle & back & frame & base & base & frame & connector wire & hose & brim \\
shoulder strap & frame & containing things & body & arm & body &  &  &  & head band & spout & button   \\
& & & & seat & & & & & & frame & panel \\
& & & & & & & & & & & bill  \\
\midrule 
 Keyboard & Knife & Lamp & Laptop & Microwave & Mug & Refrigerator & Scissors & Cabinet & Table & Trashcan & Vase \\
\midrule
 frame & handle & base & screen & body & containing things & body & blade & frame & base & base & body \\
 key & blade & body & base & base & body & base & handle & shelf & bench & cover & base \\
 & & unit & & & handle & & & drawer & tabletop & frame & containing things \\
 & & power cord & & & & & & base & & container & \\
 & & & & & & & & door & & & \\
\bottomrule
\end{tabular}
\label{tab:partnet}
\end{table}

\begin{table}[h]
\scriptsize
\renewcommand{\arraystretch}{1.0}
\renewcommand{\tabcolsep}{0.1mm}
\centering
\caption{The editable shape parameters we used to generate the geometry editing dataset and the shape properties they control. Each of the three shape categories owns an independent set of parameters.}
\begin{tabular}{llllll}
\toprule
\multicolumn{2}{c}{Chair} & \multicolumn{2}{c}{Vase} & \multicolumn{2}{c}{Table} \\
\cmidrule(lr){1-2} \cmidrule(lr){3-4} \cmidrule(lr){5-6} 
parameter & property & parameter & property & parameter & property \\
\midrule
scale\_x & width & body\_height & height & table\_top\_scale\_x & width \\
scale\_y & length & body\_width & width & table\_top\_scale\_y & length \\
scale\_z & height & bodytop\_curve\_height & bottom thickness & table\_top\_height & height \\
pillow\_state & pillow existence & body\_profile\_blend & roundness & table\_top\_shape & roundness \\
seat\_shape & roundness & handle\_count & handles existence & table\_top\_thickness & tabletop thickness \\
seat\_pos & seat height & neck\_end\_x & neck size & legs\_shape\_1 & legs thickness \\
cr\_count & back rails existence & \_end\_z & neck length & legs\_bevel & legs roundness \\
tr\_shape\_1 & back curvature & neck\_end\_bezier\_x & neck thickness & std\_legs\_bottom\_offset\_y & legs straightness \\
is\_vertical\_rail & vertical/horizontal rails &  &  & is\_std\_legs\_support\_x & leg connections existence \\
is\_back\_rest & solid back or not &  &  & std\_legs\_support\_x\_height & leg connections height \\
legs\_shape\_1 & legs thickness &  &  & legs\_support\_x\_height & leg connections thickness \\
legs\_bevel & legs roundness &  &  & is\_monoleg & one-legged or not \\
is\_monoleg & one-legged or not &  &  & is\_monoleg\_tent & leg supports existence \\
is\_monoleg\_tent & leg has supports or not &  &  & monoleg\_tent\_base\_radius & leg supports angles \\
monoleg\_bezier\_end\_x & leg thickness &  &  & monoleg\_bezier\_end\_x & leg thickness \\
frame\_top\_y\_offset\_pct & back angle &  &  &  &  \\
leg\_bottom\_y\_offset\_pct & leg angles &  &  &  &  \\
handles\_state & arms existence &  &  &  &  \\
is\_handles\_support & arm supports existence &  &  &  &  \\
handles\_profile\_width & arms thickness &  &  &  &  \\
handles\_base\_pos\_z\_pct & arms heights &  &  &  &  \\
handles\_support\_thickness & arm supports thickness &  &  &  &  \\
\bottomrule
\end{tabular}
\label{tab:geocode}
\end{table}


\section{Discussions}
\label{sec:supp_discussion}

\subsection{Point Cloud Alignment}

\begin{figure}[h]
  \centering
  \includegraphics[width=\linewidth]{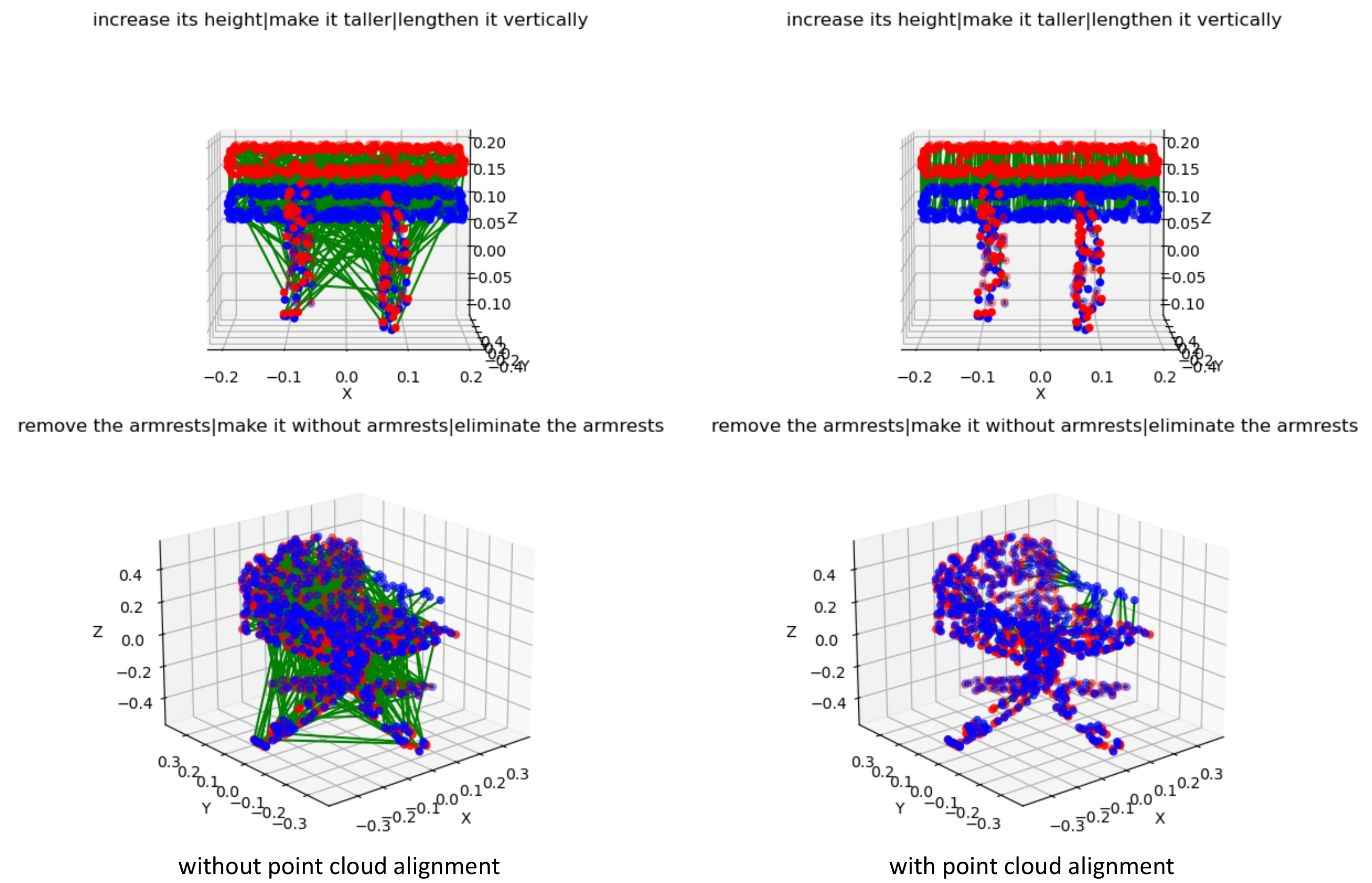}
  \caption{The effects of the point cloud alignment operation. We visualize the source and target point clouds in blue and red colors respectively, and draw the point correspondences using green lines. The edit instructions for each editing pair are annotated on the top.}
  \label{fig:alignment}
\end{figure}

As stated in Section 3.3.2, we align the source and target point clouds in the geometry editing dataset by computing a bijection $f: \mathbf{P} \rightarrow \mathbf{P}^{\prime}$ that minimizes the summarized squared distance $\sum_{j=1}^N\|\mathbf{p}_j - f(\mathbf{p}_j)\|_2^2$ before training, and then permute the target point cloud to make $\mathbf{p}_j$ and $f(\mathbf{p}_j)$ share the same order. We visualize the effects of this point cloud alignment operation in Figure~\ref{fig:alignment}. As Figure~\ref{fig:alignment} shows, the distances between the source point cloud (blue) and the target point are minimized after alignment (right colomn), while the point correspondences (green lines) are chaos and highly-random without alignment (left column). By aligning the source and target point clouds, the point offsets our model needs to learn are minimized, thus making the training easier.

\subsection{Comparison with ChangeIt3D}

ChangeIt3D~\cite{achlioptas2022changeIt3D} is the most relevant work since it is also a language-driven shape editing framework. However, its code and data are not publicly available yet thus we cannot use it as a baseline and make a comparison. In Table~\ref{tab:changeit3d}, we compare the main features of our InstructP2P and ChangeIt3D. Our InstructP2P is an end-to-end framework that can handle both color and geometry editing on point clouds, while ChangeIt3D requires a two-stage training and cannot deal with color manipulations. Besides, ChangeIt3D is trained on the ShapeTalk~\cite{achlioptas2022changeIt3D} dataset which contains limited shape categories, while our InstructP2P is fine-tuned from Point-E~\cite{nichol2022point} trained on millions of 3D models. Therefore, our method can deal with broader shape categories. Additionally, we believe their ShapeTalk dataset can also benefit our framework by providing additional geometry editing data once it is released.

\begin{table}[h]
\small
\renewcommand{\arraystretch}{1.0}
\centering
\caption{The comparison of the main features of InstructP2P and ChangeIt3D.}
\begin{tabular}{ccccc}
\toprule
Method & Color editing & Geometry editing & End-to-end & Categories \\
\midrule
ChangeIt3D & \usym{2717} & \usym{1F5F8} & \usym{2717} & limited \\
Ours & \usym{1F5F8} & \usym{1F5F8} & \usym{1F5F8} & broad \\
\bottomrule
\end{tabular}
\label{tab:changeit3d}
\end{table}

\section{Additional Qualitative Results}
\label{sec:supp_quality}

\begin{figure}[h]
  \centering
  \includegraphics[width=\linewidth]{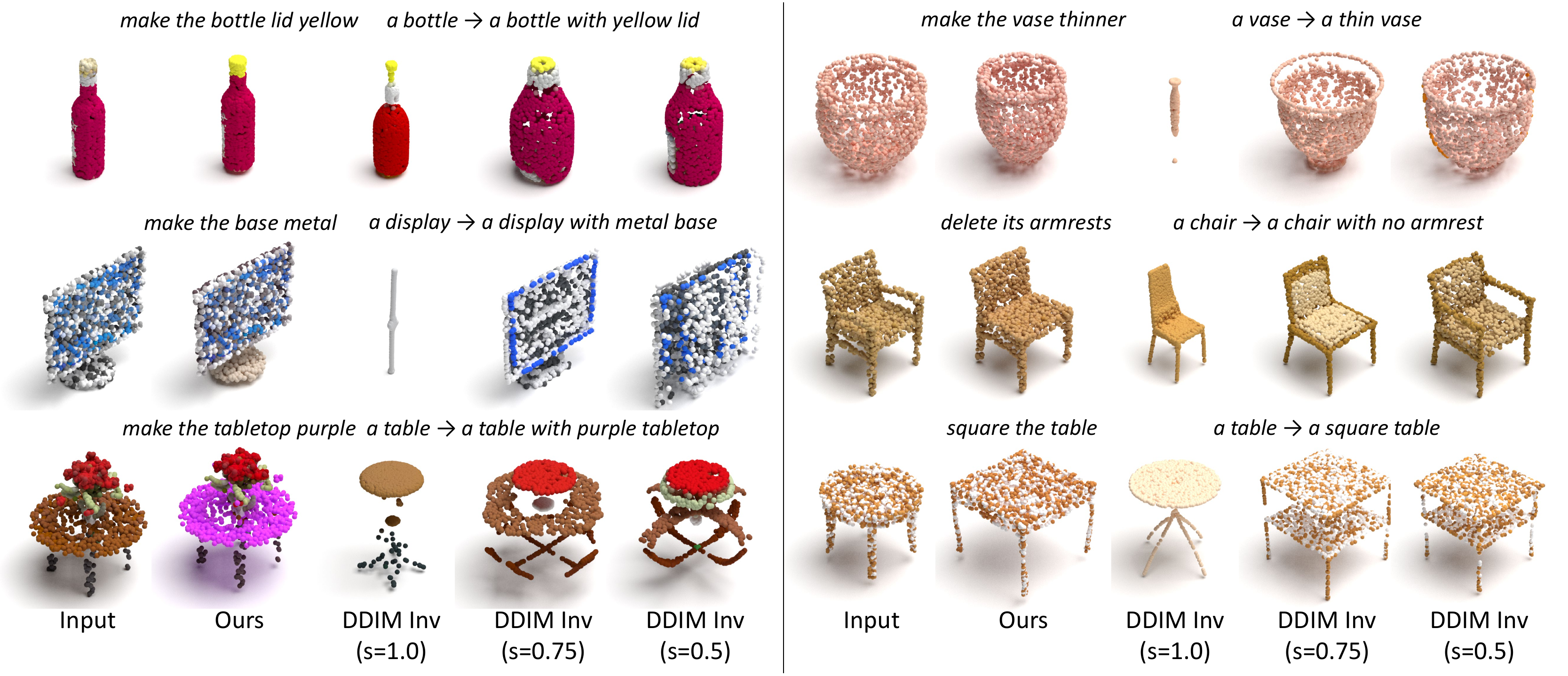}
  \vspace{-6mm}
  \caption{Additional qualitative comparisons against the DDIM inversion baseline.}
  \label{fig:results_more}
\end{figure}

\begin{figure}[h]
  \centering
  \includegraphics[width=\linewidth]{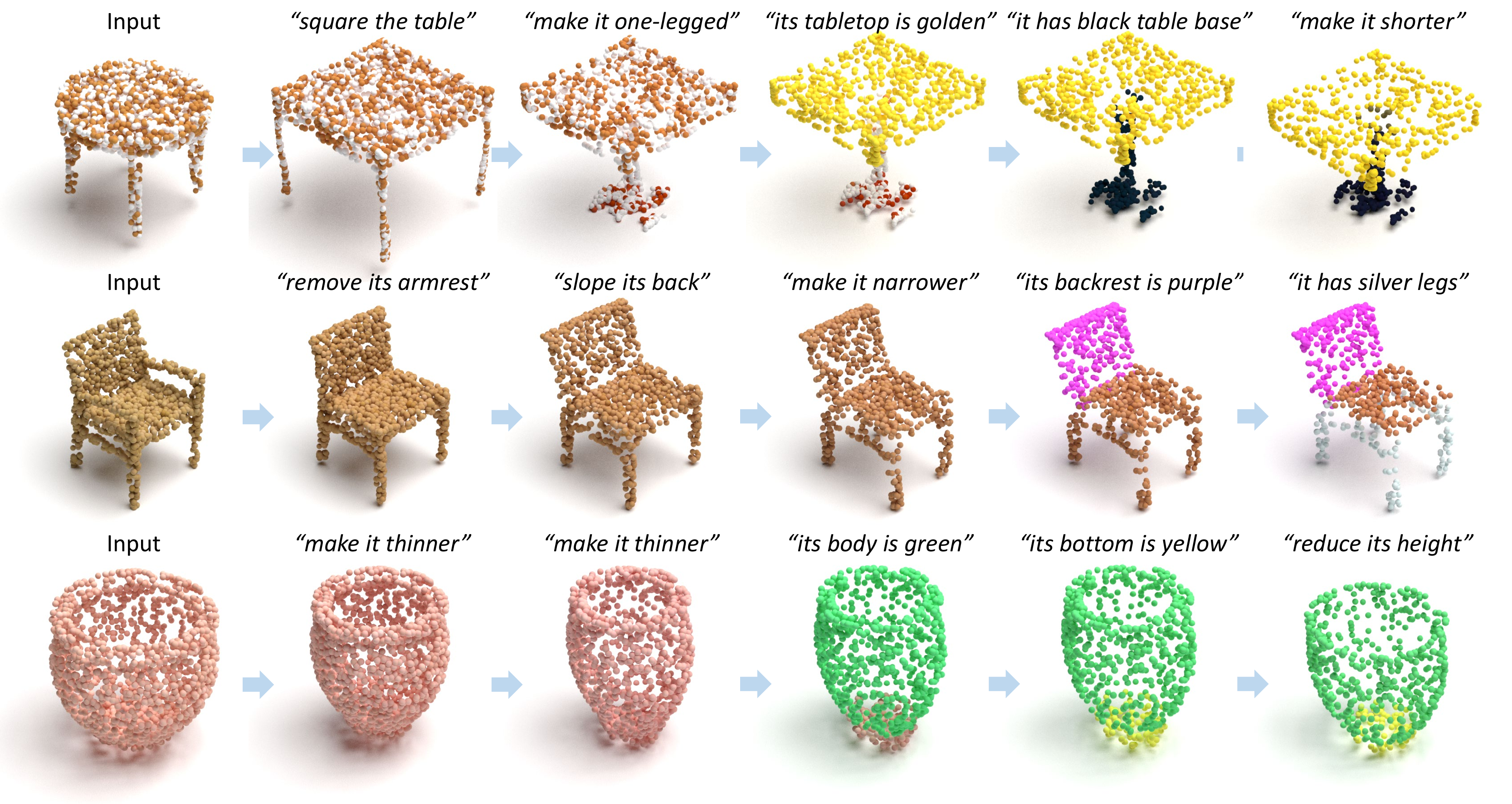}
  \vspace{-6mm}
  \caption{Additional sequential point cloud editing results.}
  \label{fig:sequential_more}
\end{figure}

\begin{figure}[h]
  \centering
  \includegraphics[width=\linewidth]{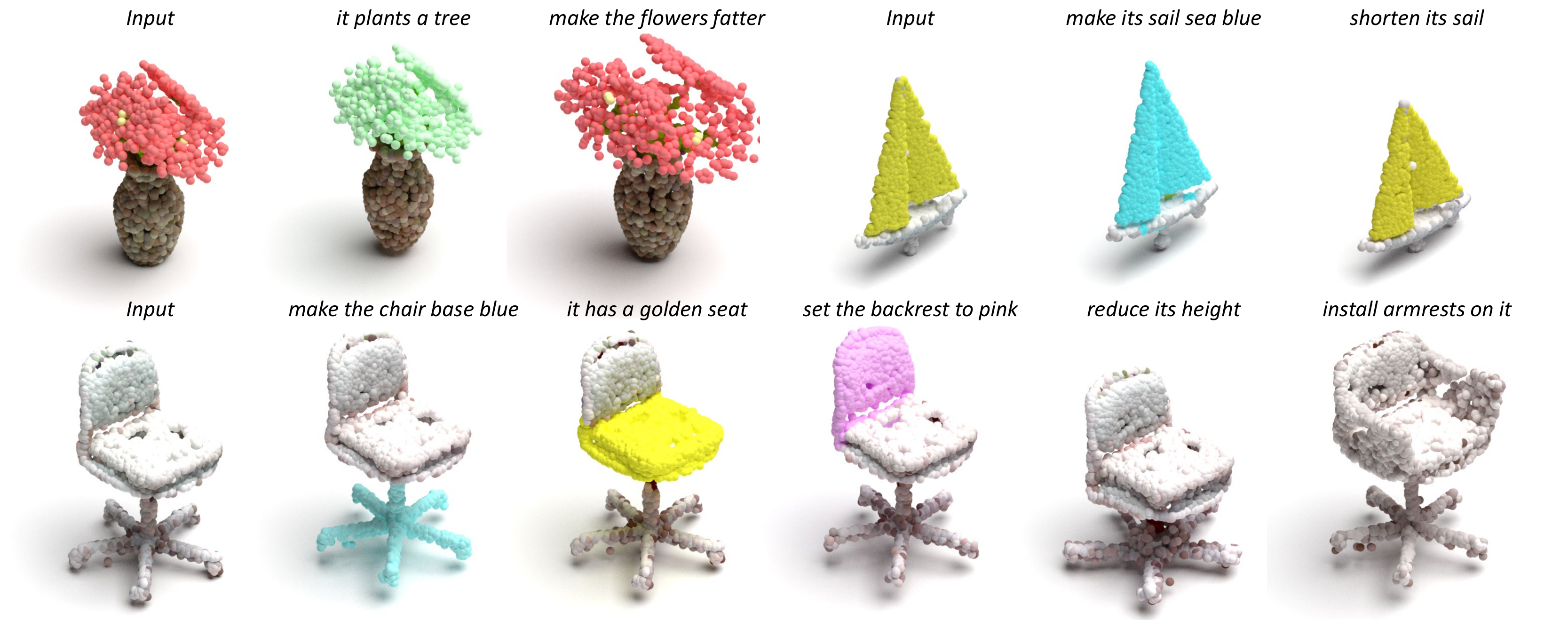}
  \vspace{-6mm}
  \caption{The generalization ability of InstructP2P. Up: generalization results on unseen edit instructions and shape categories. Down: generalization results on real scan.}
  \label{fig:generalization_more}
\end{figure}

In this section, we give additional qualitative comparisons against the DDIM~\cite{song2021denoising} inversion baseline in Figure~\ref{fig:results_more}, additional sequential point cloud editing results in Figure~\ref{fig:sequential_more}, and additional results showing the generalization ability of InstructP2P in Figure~\ref{fig:generalization_more}. 

Figure~\ref{fig:results_more} shows that our method outperforms the DDIM inversion baseline significantly. The baseline struggles to preserve the shape details irrelevant to the desired edit, even altering the strength value. The results in Figure~\ref{fig:sequential_more} demonstrate the sequentail point cloud editing ability of our method, which can be practical in real-world applications. In the first row of Figure~\ref{fig:generalization_more}, we show the generalization ability of our method to unseen edit instructions (\textit{e.g., ``it plants a tree''}) and shape categories (we take the ``ship'' category as an example). In the second row, we show the generalization ability of our method to real data. We take the real scan of a chair as input and edit it with 5 different instructions covering both color and geometry editing. Our model performs the desired edit successfully, demonstrating that InstructP2P can be applied to both synthetic and real data. 

\end{document}